%% file: main.tex
\documentclass[11pt]{article}

\usepackage{fullpage}
\usepackage[ruled, linesnumbered, vlined, commentsnumbered]{algorithm2e}
\usepackage{graphicx,subfig} 
\usepackage{amsfonts,amssymb,amsmath,amsthm,amsopn}	
\usepackage{hhline,booktabs,colortbl,multirow,tabularx,threeparttable} 
\usepackage[listings,skins,breakable]{tcolorbox}

\usepackage{enumerate}
\usepackage{authblk}
\usepackage{footnote}
\usepackage{hyperref}
\usepackage{prettyref}
\usepackage{cite}
\usepackage{setspace}
\usepackage{color}
\usepackage{xcolor}  
\usepackage{scrpage2}
\usepackage{geometry}

\usepackage{tikz}
\usepackage{pgfplots}
\usetikzlibrary{positioning,shapes,shadows,arrows,calc}
\tikzstyle{component}=[rectangle, draw=black, rounded corners, fill=blue!40, drop shadow, text centered, anchor=north, text=white, minimum height=1cm]
\tikzstyle{arrow}=[->, thick]

\pgfplotsset{compat=1.12}
\usetikzlibrary{intersections}
\usetikzlibrary{pgfplots.statistics}
\usepgfplotslibrary{fillbetween}

\def\hlinew#1{%
  \noalign{\ifnum0=`}\fi\hrule \@height #1 \futurelet
   \reserved@a\@xhline}
\makeatother

\definecolor{myblue}{RGB}{34,31,217}
\definecolor{mycyan}{gray}{.7}
\definecolor{Gray}{gray}{0.9}

\DeclareMathOperator*{\argmin}{argmin}

\newcommand{\pref}{\prettyref}

\newrefformat{fig}{Fig.~\ref{#1}}
\newrefformat{tab}{Table~\ref{#1}}
\newrefformat{sec}{Section~\ref{#1}}
\newrefformat{app}{Appendix~\ref{#1}}
\newrefformat{alg}{Algorithm~\ref{#1}}
\newrefformat{property}{Property~\ref{#1}}
\newrefformat{theorem}{Theorem~\ref{#1}}
\newrefformat{corollary}{Corollary~\ref{#1}}
\newrefformat{proposition}{Proposition~\ref{#1}}
\newrefformat{def}{Definition~\ref{#1}}
\newrefformat{eq}{equation~(\ref{#1})}

\title{\vspace{-1ex}\LARGE\textbf{Surrogate Assisted Evolutionary Algorithm for Medium Scale Expensive Multi-Objective Optimisation Problems}\footnote{This manuscript is submitted for possible publication.}}

\author[1]{\normalsize Xiaoran Ruan}
\author[2]{\normalsize Ke Li}
\author[3]{\normalsize Bilel Derbel}
\author[3]{\normalsize Arnaud Liefooghe}
\affil[1]{\normalsize College of Computer Science and Engineering, University of Electronic Science and Technology of China, 611731, Chengdu, China}
\affil[2]{\normalsize Department of Computer Science, University of Exeter, EX4 4QF, Exeter, UK}
\affil[3]{\normalsize Computer Science Department, Faculty of Science and Technology, University of Lille}
\affil[$\ast$]{\normalsize Email: \texttt{k.li@exeter.ac.uk}}

\date{}

\begin{document}
\maketitle

\vspace{-3ex}
{\normalsize\textbf{Abstract: }    Building a surrogate model of an objective function has shown to be effective to assist evolutionary algorithms (EAs) to solve real-world complex optimisation problems which involve either computationally expensive numerical simulations or costly physical experiments. However, their effectiveness mostly focuses on small-scale problems with less than 10 decision variables. The scalability of surrogate assisted EAs (SAEAs) have not been well studied yet. In this paper, we propose a Gaussian process surrogate model assisted EA for medium-scale expensive multi-objective optimisation problems with up to 50 decision variables. There are three distinctive features of our proposed SAEA. First, instead of using all decision variables in surrogate model building, we only use those correlated ones to build the surrogate model for each objective function. Second, rather than directly optimising the surrogate objective functions, the original multi-objective optimisation problem is transformed to a new one based on the surrogate models. Last but not the least, a subset selection method is developed to choose a couple of promising candidate solutions for actual objective function evaluations thus to update the training dataset. The effectiveness of our proposed algorithm is validated on benchmark problems with 10, 20, 50 variables, comparing with three state-of-the-art SAEAs.}


\input{introduction}

\input{preliminaries}

\input{algorithm}

\input{experiments}

\input{conclusion}

\section*{Acknowledgement}
K. Li was supported by UKRI Future Leaders Fellowship (Grant No. MR/S017062/1) and Royal Society (Grant No. IEC/NSFC/170243).

\bibliographystyle{IEEEtran}
\bibliography{IEEEabrv,saea}

\end{document}

%% file: introduction.tex

\section{Introduction}
\label{sec:introduction}

Multi-objective optimisation problems (MOPs) are ubiquitous in real-world applications, such as integrated circuit design~\cite{Lyu0YZ018}, water distribution network design~\cite{PierroKSB09} and aerodynamic design~\cite{EmmerichGN06}. The MOP considered in this paper is defined as:
\begin{equation}
\begin{aligned}
    &\mathrm{minimize} \quad \mathbf{F}(\mathbf{x})=\left(f_1(\mathbf{x}),\cdots,f_m(\mathbf{x})\right)^T \\
    &\mathrm{subject\ to } \quad \mathbf{x}\in\Omega
\end{aligned},
\end{equation}
where $\mathbf{x}=(x_1,\cdots,x_n)^T\in\Omega$ is a decision variable (vector), $\Omega=\Pi_{i=1}^n[a_i,b_i]\subseteq\mathbb{R}^n$ is the decision space, $\mathbf{F}:\Omega\rightarrow\mathbb{R}^m$ consists of $m$ conflicting objective functions and $\mathbb{R}^m$ is the objective space. A solution $\mathbf{x}^1\in\Omega$ is said to dominate $\mathbf{x}^2\in\Omega$, denoted as $\mathbf{x}^1\preceq\mathbf{x}^2$, if and only if $\mathbf{F}(\mathbf{x}^1)$ is not worse than $\mathbf{F}(\mathbf{x}^2)$ in any objective and it has at least one better objective. A solution $\mathbf{x}^\ast\in\Omega$ is called Pareto-optimal in case there does not exist any solution $\mathbf{x}\in\Omega$ that dominates $\mathbf{x}^\ast$. Different from global optimisation, there does not exist a global optimum that optimises all conflicting objectives. Instead, multi-objective optimisation usually seek a set of Pareto-optimal solutions, termed as Pareto-optimal set (PS), that achieve the best possible trade-off among conflicting objectives. The image of PS in the objective space is called the Pareto-optimal front (PF).

Due to the population-based property, evolutionary algorithms (EAs) have been widely applied for solving MOPs. Over the past three decades and beyond, many efforts have been devoted to the development of evolutionary multi-objective optimisation (EMO) algorithms~\cite{LiZKLW14,LiFKZ14,LiDZK15,LiDAY17,LiZ19}, such as fast non-dominated sorting genetic algorithm (NSGA-II)~\cite{DebAPM02}, indicator-based EA (IBEA)~\cite{ZitzlerK04} and multi-objective EA based on decomposition (MOEA/D)~\cite{ZhangL07}. One of the major hurdles for a wider application of EAs in real-world scenarios is their iterative nature which normally requires a vast amount of function evaluations (FEs) to approximate reasonably acceptable solution(s). This is even unacceptable in many real-world optimisation problems which involve either computationally expensive numerical simulations or costly physical experiments. For example, computational fluid dynamic simulations can take from minutes to hours to carry out one FE~\cite{JinS09}. To overcome this issue, surrogate models have shown their effectiveness to be incorporated in EAs, as known as surrogate assisted EAs (SAEAs), for solving expensive optimisation problems. However, it is worth noting that most, if not all, SAEAs are developed for small-scale problems with a relatively small number of decision variables (e.g., $n\leq 10$). As discussed in~\cite{Regis14}, the performance of SAEAs degenerate dramatically with the increase of the number of variables. This might be attributed to the defects of three major design components of a SAEA.
\vspace{-0.2em}
\begin{itemize}
    \item\textit{Surrogate modelling}: Most, if not all, surrogate models used in SAEAs are built by machine learning algorithms. It is well known that the curse-of-dimensionality is \textit{the Achilles heel} of learning algorithms. For example, the widely used Gaussian process (GP)~\cite{RasmussenW06} or Kriging model~\cite{JonesSW98} were criticised for a dramatically degenerated modelling ability with an increase of the number of variables~\cite{LiuZG14}. There have been some attempts to use other machine learning algorithms, e.g., radial-basis function networks~\cite{Regis14} and random forest~\cite{WangJ20}, to build the surrogate model for problems up to 100 variables. Another alternative solution to combat the curse-of-dimensionality is to transform the decision variables from tens of dimensions to a few dimensions by using dimensionality reduction techniques. For example, Liu et al. proposed to use Sammon mapping to enable the GP to build surrogate models in a low-dimensional space~\cite{LiuZG14}.
    \item\textit{Model-based search process}: Due to the use of surrogate model, the model-based search process is either driven towards the surrogate objective function(s) or an alternative utility function, e.g., acquisition functions used in GP assisted EAs. However, since surrogate modelling is unlikely to be accurate for high-dimensional problems, the model-based search process is highly likely to be misled. There have been some attempts to develop fine-grained search strategies to have a better exploration in the surrogate search space. For example, Sun et al.~\cite{SunJZY15} proposed a surrogate assisted cooperative particle swarm optimisation (PSO) algorithm that takes advantages of two cooperative PSO variants to balance the exploration and exploitation. In particular, the global PSO aims to identify the region(s) in which the global optimum might be located; whilst the local PSO is responsible for an intensive exploitation of those identified promising region(s). Similar idea has been studied in~\cite{YangQGJZ19} where the PSO is replaced by differential evolution. 
    \item\textit{Model management}: This step mainly aims to select promising solution(s) output from the search process for expensive objective function valuations. These newly evaluated solutions will thus be used to update the surrogate model accordingly. However, partially due to the degenerated capacity of the surrogate modelling and the model-based search process with an increase of the dimensionality, the model management becomes less effective or even pointless thus further aggravate the surrogate modelling and the search process. For example, as discussed in~\cite{TianTZSJ19}, the increase of the dimensionality makes the estimated standard deviation for measuring the uncertainty of the approximated objective function value become indifferent to each other. To overcome this issue, they proposed a multi-objective infill criterion formulation to strike a better balance between exploration and exploitation in the model management.
\end{itemize}

Bearing the above discussions in mind, this paper proposes a SAEA (dubbed SAEA/ME) for solving medium-scale expensive multi-objective optimisation problems where $n\leq 50$. In particular, we use GP to build the surrogate model given its intriguing capability to provide an estimation of not only an objective function value but also its associated uncertainty. To combat the curse-of-dimensionality, we analyse the correlation between decision variables and each objective function. Thereafter, only correlated variables for the corresponding objective function are used to build its surrogate model. During the model-based search process, the original MOP is transformed into a many-objective formulation in order to strike a balance between exploration and exploitation. In the model management step, a subset selection method is proposed to choose a couple of solutions for actual objective function evaluations and consequently to update the model. In experiments, we compare the performance of SAEA/ME with three state-of-the-art SAEAs for expensive MOPs with 10, 20 and 50 variables. Experimental results demonstrate that SAEA/ME outperforms those peer algorithms in 95 out of 108 comparisons.

The rest of this paper organised as follows. \pref{sec:preliminary} provides a pragmatic tutorial of a SAEA based on GP model which is the building block of our proposed algorithm. \pref{sec:proposed} delineates the technical details of our proposed algorithm step by step. \pref{sec:experiments} shows the empirical results along with a gentle analysis. \pref{sec:conclusion} concludes this paper and outlines some future directions.

%% file: preliminaries.tex

\section{Preliminaries}
\label{sec:preliminary}

This section provides a gentle tutorial of the working mechanism of GP surrogate model assisted EA, the pseudo-code of which is given in~\pref{alg:SAEA}. It is worth noting that \pref{alg:SAEA} is similar the efficient global optimisation (EGO)~\cite{JonesSW98} or Bayesian optimisation~\cite{ShahriariSWAF16} whilst the major difference lies in the optimiser is replaced by an EA. Although \pref{alg:SAEA} is for global optimisation, it can be generalised to the multi-objective optimisation scenario. In the following paragraphs, we elaborate on its two major components, i.e., GP regression model and acquisition functions.

\begin{algorithm}[t!]
    Use an experimental design method to sample a set of initial solutions $\mathcal{X}\leftarrow\{\mathbf{x}^i\}_{i=1}^{N_I}$ from $\Omega$ and evaluate their objective function values $\mathcal{Y}\leftarrow\{f(\mathbf{x}^i)\}_{i=1}^{N_I}$. Set the initial training dataset $\mathcal{D}\leftarrow\{(\mathbf{x}^i,f(\mathbf{x}^i)\}_{i=1}^{N_I}$;\\
\While{termination criteria is not met}{
    Build a GP model based on $\mathcal{D}$;\\
    Use EA to optimise an acquisition function to obtain a candidate solution $\mathbf{x}^\ast$;\\
    Evaluate the objective function values of $\mathbf{x}^\ast$ and set $\mathcal{D}\leftarrow\mathcal{D}\bigcup\{(\mathbf{x}^\ast,f(\mathbf{x}^\ast))\}$;
}
    \Return $\argmin\limits_{\mathbf{x}\in\mathcal{D}}f(\mathbf{x})$
    \caption{GP surrogate model assisted EA}
    \label{alg:SAEA}
\end{algorithm}

\subsection{Gaussian Process Regression Model}
\label{sec:GP}

Given a set of training data $\mathcal{D}=\{(\mathbf{x}^i,f(\mathbf{x}^i)\}_{i=1}^{N}$, GP regression model aims to learn a latent function $g(\mathbf{x})$ by assuming $f(\mathbf{x}^i)=g(\mathbf{x}^i)+\epsilon$ where $\epsilon\sim\mathcal{N}(0,\sigma^2_n)$ is an independently and identically distributed Gaussian noise. For each testing input vector $\mathbf{z}^\ast\in\Omega$, the mean and variance of the target $f(\mathbf{z}^\ast)$ are predicted as:
\begin{equation}
\begin{aligned}
\overline{g}(\mathbf{z}^\ast)&=m(\mathbf{z}^\ast)+{\mathbf{k}^\ast}^T(K+\sigma_n^2 I)^{-1}(\mathbf{f}-\mathbf{m}(X))\\
\mathbb{V}[g(\mathbf{z}^\ast)]&=k(\mathbf{z}^\ast,\mathbf{z}^\ast)-{\mathbf{k}^\ast}^T (K+\sigma_n^2 I)^{-1} {\mathbf{k}^\ast}
\end{aligned},
\label{eq:GP}
\end{equation}
where $X=(\mathbf{x}^1,\cdots,\mathbf{x}^N)^T$ and $\mathbf{f}=(f(\mathbf{x}^1),\cdots,f(\mathbf{x}^N))^T$. $\mathbf{m}(X)$ is the mean vector of $X$, $\mathbf{k}^\ast$ is the covariance vector between $X$ and $\mathbf{z}^\ast$, and $K$ is the covariance matrix of $X$. In particular, a covariance function, also known as a kernel, is used to measure the similarity between a pair of two data points $\mathbf{x}$ and $\mathbf{x}^\prime\in\Omega$. Here we use the squared exponential function in this paper and it is defined as:
\begin{equation}
k(\mathbf{x},\mathbf{x}')=\sigma_f^2\exp(-\frac{1}{2l^2}(\mathbf{x}-\mathbf{x}')^T(\mathbf{x}-\mathbf{x}')),
\end{equation}
where $\sigma_f$ is the scale parameter and $l$ is the length-scale parameter~\cite{duvenaud-thesis-2014}. Note that this covariance function is negatively related to the Euclidean distance between $\mathbf{x}$ and $\mathbf{x}'$. The predicted mean $\overline{g}(\mathbf{z}^\ast)$ is directly used as the prediction of $f(\mathbf{z}^\ast)$, and the prediction variance $\mathbb{V}[g(\mathbf{x}^\ast)]$ quantifies the uncertainty. As recommended in~\cite{RasmussenW06}, the hyperparameters associated with the mean and covariance functions are learned by maximising the log marginal likelihood function defined as:
\begin{equation}
    \begin{aligned}
        &\log p(\mathbf{f}|X)=-\frac{1}{2}(\mathbf{f}-\mathbf{m}(X))^T(K+\sigma_n^2 I)^{-1}(\mathbf{f}-\mathbf{m}(X))\\
                             &-\frac{1}{2}\log|K+\sigma_n^2 I|-\frac{N}{2}\log 2\pi
    \end{aligned}.
\end{equation}

\subsection{Acquisition Functions}
\label{sec:acq}

Instead of optimising the surrogate objective function, the search process of the GP surrogate model assisted EA considered in this paper is driven by the acquisition function. Generally speaking, an acquisition function is used to measure the value that would be generated by evaluating the objective function at a new sample point $\mathbf{x}$, based on the current posterior distribution over $f(\mathbf{x})$. There are three most popular acquisition functions in the literature. 
\begin{itemize}
    \item Probability of improvement (PI)~\cite{Kushner64}:
        \begin{equation}
            \text{PI}(\mathbf{x}) = \Phi\left(\frac{\overline{g}(\mathbf{x})-f(\mathbf{x}^\ast)}{\mathbb{V}[g(\mathbf{x})]}\right),
        \end{equation}
        where $\Phi(\cdot)$ is the cumulative distribution function (CDF) of the standard normal distribution. The PI aims to measure the probability of achieving any improvement over the current best sample point $\mathbf{x}^\ast$.
    \item Expected improvement (EI)~\cite{Jones01}:
        \begin{equation}
            \begin{aligned}
                \text{EI}(\mathbf{x}) 
                &= \mathbb{E}\left[\max(f(\mathbf{x}^\ast)-f(\mathbf{x}),0]\right] \\
                &= (f(\mathbf{x}^\ast)-\overline{g}(\mathbf{x}))\Phi\left(\frac{f(\mathbf{x}^\ast)-\overline{g}(\mathbf{x})}{\mathbb{V}[g(\mathbf{x})]}\right)\\
                &+\mathbb{V}[g(\mathbf{x})]\phi\left(\frac{f(\mathbf{x}^\ast)-\overline{g}(\mathbf{x})}{\mathbb{V}[g(\mathbf{x})]}\right)
            \end{aligned}
        \end{equation}
        where $\phi(\cdot)$ is the probability density function (PDF) of the standard normal distribution. The EI is able to evaluate the expectation of improvement over $\mathbf{x}^\ast$.
    \item Upper confidence bound (UCB)~\cite{CoxJ97}:
        \begin{equation}
            \text{UCB}(\mathbf{x})=\overline{g}(\mathbf{x})+k\mathbb{V}[g(\mathbf{x})]
        \end{equation}
        where $k$ is a control parameter used to characterise the trade-off between exploration and exploitation. 
\end{itemize}

Noting that all these acquisition functions are designed as a combination of the predicted mean and its associated variance. In this case, they all imply an aggregated way to strike a balance between exploration and exploitation. As discussed in a recent study~\cite{WilsonHD18}, optimising an acquisition function is highly multi-modal and is far from trivial.

%% file: algorithm.tex

\section{Proposed Algorithm}
\label{sec:proposed}

\begin{algorithm}[htbp]
	Use an experimental design method to sample a set of initial solutions $\mathcal{X}\leftarrow\{\mathbf{x}^i\}_{i=1}^{N_I}$ from $\Omega$ and evaluate their objective function values $\mathcal{Y}\leftarrow\{\mathbf{F}(\mathbf{x}^i)\}_{i=1}^{N_I}$. Set the initial training dataset $\mathcal{D}\leftarrow\{(\mathbf{x}^i,\mathbf{F}(\mathbf{x}^i)\}_{i=1}^{N_I}$;\\
    $\{\mathcal{G}_i\}_{i=1}^m\leftarrow$\texttt{CorrelationAnalysis}();\\
    \While{not terminated}{
    	Build GP models for each objective function based on $\mathcal{D}$;\\
		Use NSGA-II to optimise the problem shown in~\pref{eq:transformed} and output a set of solutions $\mathcal{S}$;\\    
        $\mathcal{S}^\ast\leftarrow$\texttt{SubsetSelection}($\mathcal{S}$);\\
        Evaluate the objective function values of $\mathcal{S}^\ast$ and set $\mathcal{D}\leftarrow\mathcal{D}\bigcup\{(\mathbf{x}^\ast,\mathbf{F}(\mathbf{x}^\ast))|\mathbf{x}^\ast\in\mathcal{S}^\ast\}$;
    }
    \Return all non-dominated solutions in $\mathcal{D}$
    \caption{SAEA/ME}
    \label{alg:SAEAME}
\end{algorithm}

\pref{alg:SAEAME} gives the pseudo-code of our proposed SAEA/ME. Note that we build a GP surrogate model for each objective as we are going to solve MOPs. The general framework is similar to that of \pref{alg:SAEA} whilst there are three distinctive features: 1) to reduce the dimensionality of the feature space when building a GP model, we analyse the association relationship between decision variables and each objective functions at the outset of SAEA/ME; 2) the model-based search process is driven towards a transformed many-objective optimisation problem; and 3) a subset selection mechanism is proposed to choose a couple of promising solutions for function evaluations and model management. We will elaborate on these three design components in the following paragraphs.

\subsection{Identifying Correlation Relationship between Variables and Objective Functions}
\label{sec:correlation}

As discussed in~\pref{sec:introduction}, the curse-of-the-dimensionality is one of most important reasons that leads to the degenerated surrogate modelling performance of GP with an increase of the dimensionality. Due to the existence of more than one objective function, it is highly likely that not every decision variable is correlated with each objective function. For example, the widely used test problem instance ZDT1~\cite{ZitzlerDT00} is formulated as:
\begin{equation}
    \begin{aligned}
        &f_{1}(\mathbf{x})=x_{1}\\
        &f_{2}(\mathbf{x})=g(\mathbf{x})[1-\sqrt{f_1(\mathbf{x})/g(\mathbf{x})}]
    \end{aligned},
    \label{eq:zdt1}
\end{equation}
where $g(\mathbf{x})=1+9\left(\sum_{i=2}^{n} x_{i}\right)/(n-1)$ and $\mathbf{x}=(x_1,\cdots,x_n)^T \in [0,1]^n$. According to~\pref{eq:zdt1}, it is obvious that the first objective function $f_1(\mathbf{x})$ only depends on $x_1$. In this case, the GP surrogate model of $f_1(\mathbf{x})$ can be directly built upon $x_1$ without compromising any accuracy. By doing so, we can expect a significantly reduced feature space thus leading to a mitigation of the curse-of-dimensionality. Bearing this consideration in mind, we propose to analyse the correlation relationship between decision variables and each objective function before building GP surrogate models. Inspired by the variable grouping idea proposed for problem decomposition in large-scale global optimisation~\cite{OmidvarLMY14}, the correlation analysis used in SAEA/ME is given in~\pref{alg:analysis}.

\begin{algorithm}[t!]
	\KwOut{$\{\mathcal{G}_i\}_{i=1}^m$ correlation groups between decision variables and objective functions}
	\For{$i\leftarrow 1$ \KwTo $n$}{
        $\mathcal{G}_i\leftarrow\emptyset$;
    }
    \For{$i\leftarrow 1$ \KwTo $n$}{
        $x^s_i\leftarrow a_i$;
    }
    Evaluate the objective function of $\mathbf{x}^l$;\\
    \For{$i\leftarrow 1$ \KwTo $n$}{
        $\mathbf{x}^t\leftarrow\mathbf{x}^s$, $x^t_i\leftarrow b_i$;\\
        Evaluate the objective function of $\mathbf{x}^t$;\\
        \For{$j\leftarrow 1$ \KwTo $m$}{
            $\Delta\leftarrow |f_j(\mathbf{x}^t)-f_j(\mathbf{x}^s)|$;\\
            \If{$\Delta<\delta$}{
                $\mathcal{G}_j\leftarrow\mathcal{G}_j\bigcup\{i\}$;
            }
        }
    }
    \Return $\{\mathcal{G}_i\}_{i=1}^m$
    \caption{\texttt{CorrelationAnalysis}()}
    \label{alg:analysis}
\end{algorithm}

Specifically, we first initialise a set of correlation groups for each objective function where $\mathcal{G}_i$ consists of the indices of variables correlated with the $i$-th objective function (lines 1 and 2 of~\pref{alg:analysis}). Thereafter, we generate a sentinel solution $\mathbf{x}^s$ whose decision variables are all set to be the lower bounds (i.e., $a_i$, $i\in\{1,\cdots,n\}$) of the underlying MOP (lines 3 and 4 of~\pref{alg:analysis}). During the main for-loop, each variable of $\mathbf{x}^s$ is perturbed to the upper bound (i.e., $b_i$, $i\in\{1,\cdots,n\}$) of the underlying MOP (lines 7 and 8 of~\pref{alg:analysis}). If we observe a \textit{significant} change at the $j$-th objective between $\mathbf{x}^s$ and its perturbed solution, the $i$-th variable is thus considered to be correlated with the $j$-th objective (lines 9 to 12 of~\pref{alg:analysis}). In particular, the significance level $\delta>0$ is set to be a small number which is set as $10^{-6}$ in this paper. Different from the grouping operation in~\cite{OmidvarLMY14}, which requires $\mathcal{O}(n^2)$ function evaluations, the correlation analysis in~\pref{alg:analysis} only requires $n+1$ function evaluations which is acceptable even under a computationally expensive optimisation scenario.

\subsection{Search Based on a Transformed Many-Objective Optimisation Problem}
\label{sec:problem_form}

As discussed in~\pref{sec:introduction}, the model-based search process can be misled, especially when having a large number of variables, by an inappropriate problem formulation. In particular, the widely used acquisition functions, as discussed in~\pref{sec:acq}, in many GP model assisted EAs are essentially linear combinations of exploration and exploitation. As reported in~\cite{Berger-TalNMS14}, there exist natural trade-offs between exploration and exploitation, a linear combination between which does not fully reflect their trade-offs. In this paper, we propose a transformation of the original $m$-objective problem into a $2m$-objective problem formation as follows: 
\begin{equation}
    \begin{array}{ll}
        \mathrm{minimize} & {\mathbf{H}(\mathbf{x})=\left(h_{1}(\mathbf{x}),\cdots,h_{2m}(\mathbf{x})\right)^{T}} \\ 
        \mathrm{subject\ to } & \mathbf{x}\in\Omega
    \end{array},
    \label{eq:transformed}
\end{equation}
where $h_{2i-1}(\mathbf{x})=\overline{g}_i(\mathbf{x})$ and $h_{2i}(\mathbf{x})=\overline{g}_i(\mathbf{x})-\mathbb{V}[g_i(\mathbf{x})]$, $i\in\{1,\cdots,m\}$. In this sense, each of those original objective functions is transformed into another two surrogate objective functions: one is the predicted mean whilst the other is used to evaluate the uncertain over the prediction. In particular, the uncertainty term gives a lower confidence bound at the predicted point. Here we use NSGA-II as the EA to optimise the newly formed optimisation problem.

\subsection{Model Management Based on Subset Selection}
\label{sec:subset}

\begin{figure}
\centering
\includegraphics[width=0.4\textwidth]{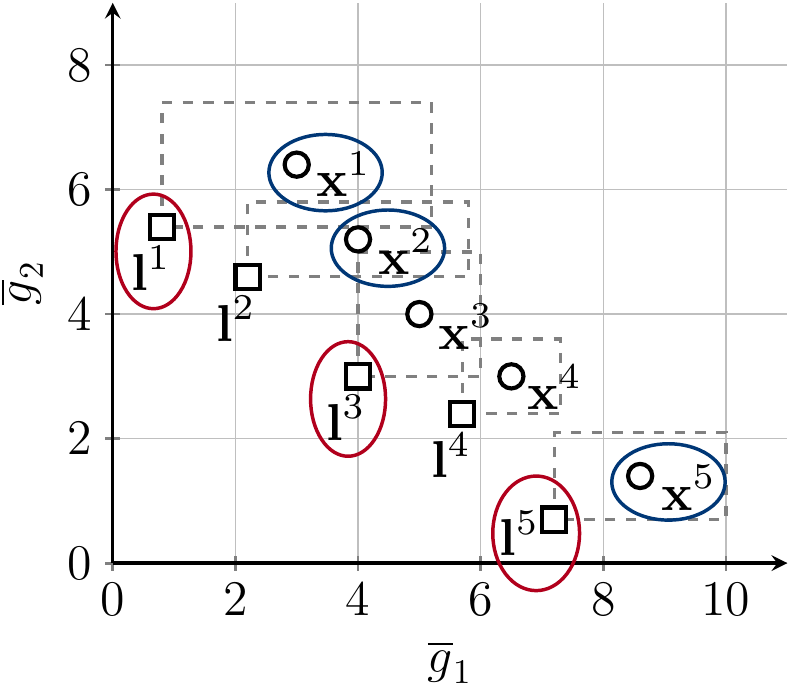}
\caption{An illustrative example our subset selection method for model management.}
\label{fig:selectSOI}
\end{figure}

\begin{algorithm}[t!]
	\KwIn{$\mathcal{S}$ solutions output by the search process}
	\KwOut{$\mathcal{S}^k$ solutions for model management}
    Use the GP model to predict the objective function values and their variances in $\mathcal{S}$ and obtain $\mathcal{S}^o\leftarrow\{\overline{\mathbf{g}}(\mathbf{x})|\mathbf{x}\in\mathcal{S}\}$;\\
    Set $\mathcal{S}^l\leftarrow\{\mathbf{l}|l_i=\overline{g}_i(\mathbf{x})-2\mathbb{V}[g_i(\mathbf{x})],i=1,\cdots,m,\mathbf{x}\in\mathcal{S}\}$;\\
    \For{$i\leftarrow 1$\KwTo $|\mathcal{S}|$}{
        \tcc*[h]{\textcolor{green!60!black!80!}{$\mathtt{HV}(\mathcal{S})$ is the HV of $\mathcal{S}$ and $\mathtt{HVC}(\mathbf{x})$ is the HV contribution of $\mathbf{x}$}}
        $\mathtt{HVC}(\mathbf{x}^i)\leftarrow \mathtt{HV}(\mathcal{S})-\mathtt{HV}(\mathcal{S}\setminus\{\mathbf{x}^i\})$;\\
        $\mathtt{HVC}(\mathbf{l}^i)\leftarrow \mathtt{HV}(\mathcal{S}^l)-\mathtt{HV}(\mathcal{S}^l\setminus\{\mathbf{l}^i\})$;\\
    }
    Sort the HV contributions of each solution in $\mathcal{S}$ and store the top $k$ solutions in $\mathcal{S}^k_o$;\\
    Sort the HV contributions of each solution in $\mathcal{S}^l$ and store the top $k$ solutions in $\mathcal{S}^k_l$;\\

    \Return $\mathcal{S}^k\leftarrow\mathcal{S}^k_o\bigcup\mathcal{S}^k_l$

    \caption{\texttt{SubsetSelection($\mathcal{S})$}}
    \label{alg:subset_selection}
\end{algorithm}

As discussed in~\pref{sec:introduction}, the major purpose of the model management is to select promising solution(s) for actual function evaluation(s) thus to update the model. Classic GP model assisted EAs were designed to carry out the model management in a sequential manner where only one solution is selected for the actual function evaluation at a time. As discussed in~\cite{GonzalezDHL16}, it is desirable to simultaneously evaluate multiple solutions in a batch manner thus to facilitate a better parallelism. Bearing this consideration in mind, this paper proposes a subset selection method to choose a couple of solutions for function evaluations. In particular, subset selection is a post-hoc method that is able to find a pre-defined number of solutions from a population for Hypervolume (HV) maximisation~\cite{IshibuchiISN17}. As a result, these selected solutions are representative enough to resemble the PF for performance benchmarking. However, since the solutions returned by the model-based search process are evaluated by surrogate models, the subset selection merely based on the predicted objective function values is inevitably error-prone. To mitigate the uncertainty brought by the surrogate models, our subset selection for model management also takes the estimated variance into account.

The pseudo-code of our proposed subset selection method is given in~\pref{alg:subset_selection}. To have a better intuition, let us explain \pref{alg:subset_selection} by an illustrative example shown in~\pref{fig:selectSOI}. Suppose that there are five solutions $\mathcal{S}=\{\mathbf{x}^i\}_{i=1}^5$ returned by the model-based search process. According the GP model in~\pref{eq:GP}, their predicted mean objective functions can be predicted and constitute $\mathcal{S}^o=\{\overline{\mathbf{g}}(\mathbf{x})|\mathbf{x}\in\mathcal{S}\}_{i=1}^5$ (line 1 of~\pref{alg:subset_selection}). By taking the estimated variance into account, each $\overline{\mathbf{g}}(\mathbf{x})$ is surrounded by a rectangle which represents a 95\%$\times$95\% confidence level. Accordingly, its corresponding rectangle is bounded by $\overline{g}_i(\mathbf{x})\pm2\mathbb{V}[g_i(\mathbf{x})]$ at the $i$-th objective. It is worth noting that such derivation can be easily generalised to more than two-dimensional space. Since the transformed many-objective optimisation formulation used in the model-based search process takes the lower confidence bound into consideration, here we are only interested in the lower confidence bounds (i.e., those lower bound vertex of each rectangle $\mathcal{S}^l=\{\mathbf{l}^i|l^i_j=\overline{g}_j(\mathbf{x})-2\mathbb{V}[g_j(\mathbf{x})]$, $j\in\{1,\cdots,m\}\}_{i=1}^5$, line 2 of~\pref{alg:subset_selection}). During our subset selection process, we use the Hypervolume (HV) contribution as the criterion to choose the top $k\geq 1$ solutions from both $\mathcal{S}^o$ and $\mathcal{S}^l$ (lines 3 to 7 of~\pref{alg:subset_selection}). Note that $k$ is a control parameter and it is set as 10 in our experiments. At the end, the intersection between $\mathcal{S}^o$ and $\mathcal{S}^l$ is used for actual function evaluations. Let us look back to the illustrative example shown in~\pref{fig:selectSOI}, by setting $k=3$, we have the solutions selected from $\mathcal{S}^o$ are $\{\mathbf{x}^2,\mathbf{x}^3,\mathbf{x}^4\}$ whilst those selected from $\mathcal{S}^l$ are $\{\mathbf{l}^1,\mathbf{l}^3,\mathbf{l}^4\}$. Finally, only $\mathbf{x}^3$ and $\mathbf{x}^4$ are chosen for actual function evaluations and model management thereafter.

%% file: experiments.tex

\section{Empirical Study}
\label{sec:experiments}

\begin{table*}[htbp]
	\scriptsize
    \centering
    \begin{threeparttable}
        \caption{Comparison Results of SAEA/ME Against Three State-of-the-art SAEAs on 12 Test Problems with 10, 20 and 50 Decision Variables.}
        \begin{tabular}{c|c|c|c|c|c}
            \hline
            & $d$     & ParEGO & MOEA/D-EGO & K-RVEA & SAEA/ME \\
            \hline
            \multirow{3}[2]{*}{ZDT1} & 10    & \cellcolor[rgb]{ .749,  .749,  .749}\textbf{2.376E-2 (7.20E-3)}$^{\ddagger}$ & 1.634E-1 (1.46E-1) & 2.702E-2 (6.82E-3)$^{\ddagger}$ & 1.290E-1 (7.32E-2) \\
            & 20    & 1.554E-1 (3.78E-2)$^{\dagger}$ & 3.031E+1 (9.46E+0)$^{\dagger}$ & 1.083E-1 (5.98E-2)$^{\dagger}$ & \cellcolor[rgb]{ .749,  .749,  .749}\textbf{2.847E-2 (1.93E-2)} \\
            & 50    & 1.666E+1 (7.00E+0)$^{\dagger}$ & 1.120E+2 (2.73E+1)$^{\dagger}$ & 1.529E+2 (2.90E+1)$^{\dagger}$ & \cellcolor[rgb]{ .749,  .749,  .749}\textbf{9.662E-3 (2.31E-3)} \\
            \hline
            \hline
            \multirow{3}[2]{*}{ZDT2} & 10    & \cellcolor[rgb]{ .749,  .749,  .749}\textbf{2.211E-2 (4.27E-3)}$^{\ddagger}$ & 2.488E-1 (1.14E-1)$^{\dagger}$ & 4.866E-2 (4.72E-2) & 4.055E-2 (1.74E-2) \\
            & 20    & 1.919E-1 (7.78E-2)$^{\dagger}$ & 2.804E+1 (1.24E+1)$^{\dagger}$ & 2.870E-1 (1.39E-1)$^{\dagger}$ & \cellcolor[rgb]{ .749,  .749,  .749}\textbf{5.641E-2 (3.33E-2)} \\
            & 50    & 1.143E+1 (2.96E+0)$^{\dagger}$ & 1.145E+2 (3.04E+1)$^{\dagger}$ & 1.632E+2 (6.23E+0)$^{\dagger}$ & \cellcolor[rgb]{ .749,  .749,  .749}\textbf{1.659E-2 (8.33E-3)} \\
            \hline
            \hline
            \multirow{3}[2]{*}{ZDT3} & 10    & 7.567E-2 (2.25E-2)$^{\ddagger}$ & 3.979E-1 (1.83E-1)$^{\dagger}$ & \cellcolor[rgb]{ .749,  .749,  .749}\textbf{5.917E-2 (3.45E-2)}$^{\ddagger}$ & 1.121E-1 (5.81E-2) \\
            & 20    & 1.857E-1 (3.71E-2)$^{\dagger}$ & 2.333E+1 (9.79E+0)$^{\dagger}$ & \cellcolor[rgb]{ .749,  .749,  .749}\textbf{1.392E-1 (8.08E-2)} & 1.424E-1 (1.02E-1) \\
            & 50    & 1.691E+1 (6.70E+0)$^{\dagger}$ & 1.010E+2 (2.99E+1)$^{\dagger}$ & 1.576E+2 (5.42E+0)$^{\dagger}$ & \cellcolor[rgb]{ .749,  .749,  .749}\textbf{8.997E-2 (1.09E-1)} \\
            \hline
            \hline
            \multirow{3}[2]{*}{ZDT4} & 10    & 7.124E+1 (1.13E+1)$^{\dagger}$ & 7.892E+1 (1.18E+1)$^{\dagger}$ & \cellcolor[rgb]{ .749,  .749,  .749}\textbf{2.996E+1 (1.39E+1)}$^{\ddagger}$ & 5.398E+1 (1.8E+1) \\
            & 20    & 1.830E+2 (2.27E+1)$^{\dagger}$ & 2.167E+2 (1.49E+1)$^{\dagger}$ & \cellcolor[rgb]{ .749,  .749,  .749}\textbf{1.269E+2 (2.92E+1)}$^{\ddagger}$ & 1.394E+2 (2.11E+1) \\
            & 50    & 6.280E+2 (4.59E+1)$^{\dagger}$ & 6.431E+2 (3.88E+1)$^{\dagger}$ & 6.369E+2 (4.05E+1)$^{\dagger}$ & \cellcolor[rgb]{ .749,  .749,  .749}\textbf{3.997E+2 (2.52E+1)} \\
            \hline
            \hline
            \multirow{3}[2]{*}{ZDT6} & 10    & 4.221E-1 (9.97E-2)$^{\dagger}$ & 1.233E+0 (1.23E+0)$^{\dagger}$ & 1.467E+0 (2.48E-1)$^{\dagger}$ & \cellcolor[rgb]{ .749,  .749,  .749}\textbf{1.188E-1 (5.51E-2)} \\
            & 20    & 3.801E+0 (6.53E-1)$^{\dagger}$ & 1.245E+1 (1.52E+0)$^{\dagger}$ & 3.101E+0 (6.63E-1)$^{\dagger}$ & \cellcolor[rgb]{ .749,  .749,  .749}\textbf{9.490E-2 (2.42E-2)} \\
            & 50    & 1.295E+1 (8.20E-1)$^{\dagger}$ & 1.836E+1 (6.48E-1)$^{\dagger}$ & 1.276E+1 (4.17E+0)$^{\dagger}$ & \cellcolor[rgb]{ .749,  .749,  .749}\textbf{6.603E-2 (2.56E-2)} \\
            \hline
            \hline
            \multirow{3}[2]{*}{DTLZ1} & 10    & \cellcolor[rgb]{ .749,  .749,  .749}\textbf{6.295E+1 (7.08E+0)}$^{\ddagger}$ & 8.388E+1 (1.21E+1)$^{\ddagger}$ & 8.394E+1 (1.92E+1)$^{\ddagger}$ & 9.765E+1 (1.43E+1) \\
            & 20    & \cellcolor[rgb]{ .749,  .749,  .749}\textbf{2.346E+2 (1.51E+1)}$^{\ddagger}$ & 2.752E+2 (6.31E+1) & 3.102E+2 (5.17E+1) & 3.062E+2 (4.69E+1) \\
            & 50    & 1.193E+3 (5.67E+1)$^{\dagger}$ & 1.063E+3 (2.14E+2) & 1.142E+3 (7.67E+1)$^{\dagger}$ & \cellcolor[rgb]{ .749,  .749,  .749}\textbf{1.032E+3 (1.05E+2)} \\
            \hline
            \hline
            \multirow{3}[2]{*}{DTLZ2} & 10    & 3.665E-1 (3.65E-2)$^{\dagger}$ & 3.307E-1 (3.05E-2)$^{\dagger}$ & 1.244E-1 (1.41E-2)$^{\dagger}$ & \cellcolor[rgb]{ .749,  .749,  .749}\textbf{8.815E-2 (5.34E-3)} \\
            & 20    & 8.518E-1 (7.37E-2)$^{\dagger}$ & 6.444E-1 (8.22E-2)$^{\dagger}$ & 4.812E-1 (5.67E-2)$^{\dagger}$ & \cellcolor[rgb]{ .749,  .749,  .749}\textbf{1.231E-1 (8.56E-3)} \\
            & 50    & 2.677E+0 (9.67E-2)$^{\dagger}$ & 2.004E+0 (2.12E-1)$^{\dagger}$ & 2.079E+0 (1.20E-1)$^{\dagger}$ & \cellcolor[rgb]{ .749,  .749,  .749}\textbf{3.542E-1 (9.96E-2)} \\
            \hline
            \hline
            \multirow{3}[2]{*}{DTLZ3} & 10    & \cellcolor[rgb]{ .749,  .749,  .749}\textbf{1.721E+2 (1.05E+1)}$^{\ddagger}$ & 1.984E+2 (2.61E+1)$^{\ddagger}$ & 2.252E+2 (6.87E+1) & 2.452E+2 (5.63E+1) \\
            & 20    & \cellcolor[rgb]{ .749,  .749,  .749}\textbf{4.839E+2 (5.72E+1)}$^{\ddagger}$ & 5.837E+2 (1.56E+2)$^{\ddagger}$ & 8.493E+2 (1.49E+2) & 8.539E+2 (1.44E+2) \\
            & 50    & 3.666E+3 (2.01E+2)$^{\dagger}$ & 2.979E+3 (7.18E+2)$^{\dagger}$ & 3.387E+3 (1.60E+2)$^{\dagger}$ & \cellcolor[rgb]{ .749,  .749,  .749}\textbf{2.127E+3 (5.59E+2)} \\
            \hline
            \hline
            \multirow{3}[2]{*}{DTLZ4} & 10    & 6.286E-1 (1.02E-1) & 6.363E-1 (5.18E-2) & \cellcolor[rgb]{ .749,  .749,  .749}\textbf{3.076E-1 (9.98E-2)}$^{\ddagger}$ & 6.377E-1 (1.41E-1) \\
            & 20    & 1.108E+0 (2.22E-1)$^{\dagger}$ & 1.165E+0 (1.15E-1)$^{\dagger}$ & \cellcolor[rgb]{ .749,  .749,  .749}\textbf{8.426E-1 (1.49E-1)}$^{\ddagger}$ & 9.564E-1 (1.05E-1) \\
            & 50    & 2.859E+0 (3.74E-1)$^{\dagger}$ & 2.431E+0 (2.66E-1)$^{\dagger}$ & 3.148E+0 (9.81E-2)$^{\dagger}$ & \cellcolor[rgb]{ .749,  .749,  .749}\textbf{1.139E+0 (1.17E-1)} \\
            \hline
            \hline
            \multirow{3}[2]{*}{DTLZ5} & 10    & 2.711E-1 (4.09E-2)$^{\dagger}$ & 2.569E-1 (3.23E-2)$^{\dagger}$ & 7.247E-2 (1.12E-2)$^{\dagger}$ & \cellcolor[rgb]{ .749,  .749,  .749}\textbf{3.890E-2 (1.86E-2)} \\
            & 20    & 7.485E-1 (1.02E-1)$^{\dagger}$ & 5.330E-1 (5.80E-2)$^{\dagger}$ & 4.157E-1 (7.30E-2)$^{\dagger}$ & \cellcolor[rgb]{ .749,  .749,  .749}\textbf{7.247E-2 (1.32E-2)} \\
            & 50    & 2.627E+0 (1.14E-1)$^{\dagger}$ & 1.917E+0 (3.71E-1)$^{\dagger}$ & 1.992E+0 (3.22E-1)$^{\dagger}$ & \cellcolor[rgb]{ .749,  .749,  .749}\textbf{2.581E-1 (4.35E-2)} \\
            \hline
            \hline
            \multirow{3}[2]{*}{DTLZ6} & 10    & 1.164E+0 (3.56E-1)$^{\dagger}$ & 1.887E+0 (7.81E-1)$^{\dagger}$ & 2.902E+0 (3.60E-1)$^{\dagger}$ & \cellcolor[rgb]{ .749,  .749,  .749}\textbf{5.782E-1 (2.74E-1)} \\
            & 20    & 7.287E+0 (7.85E-1)$^{\dagger}$ & 6.757E+0 (1.54E+0)$^{\dagger}$ & 1.021E+1 (7.42E-1)$^{\dagger}$ & \cellcolor[rgb]{ .749,  .749,  .749}\textbf{1.939E+0 (9.58E-1)} \\
            & 50    & 3.657E+1 (6.82E-1)$^{\dagger}$ & 2.723E+1 (3.93E+0)$^{\dagger}$ & 3.848E+1 (1.24E+0)$^{\dagger}$ & \cellcolor[rgb]{ .749,  .749,  .749}\textbf{2.073E+1 (5.59E+0)} \\
            \hline
            \hline
            \multirow{3}[2]{*}{DTLZ7} & 10    & 1.782E-1 (2.14E-2) & 2.284E-1 (6.70E-2)$^{\dagger}$ & \cellcolor[rgb]{ .749,  .749,  .749}\textbf{1.105E-1 (1.07E-2)}$^{\ddagger}$ & 1.909E-1 (7.33E-2) \\
            & 20    & 2.177E-1 (4.13E-2)$^{\dagger}$ & 3.829E+0 (2.01E+0)$^{\dagger}$ & 2.783E-1 (1.97E-1)$^{\dagger}$ & \cellcolor[rgb]{ .749,  .749,  .749}\textbf{1.130E-1 (3.93E-2)} \\
            & 50    & 1.823E+0 (3.97E-1)$^{\dagger}$ & 8.246E+0 (1.06E+0)$^{\dagger}$ & 6.520E+0 (3.88E+0)$^{\dagger}$ & \cellcolor[rgb]{ .749,  .749,  .749}\textbf{7.334E-2 (4.52E-3)} \\
            \hline
        \end{tabular}%
        \begin{tablenotes}
        \item $^\dagger$ denotes that the better IGD value obtained by SAEA/ME is significantly better than the corresponding peer algorithm according to the Wilcoxon's rank sum test at a 5\% significance level; whilst $^\ddagger$ denotes an opposite conclusion. 
        \end{tablenotes}
        \label{tab:results}%
    \end{threeparttable}
\end{table*}%

\begin{figure*}
    \includegraphics[width=.22\textwidth]{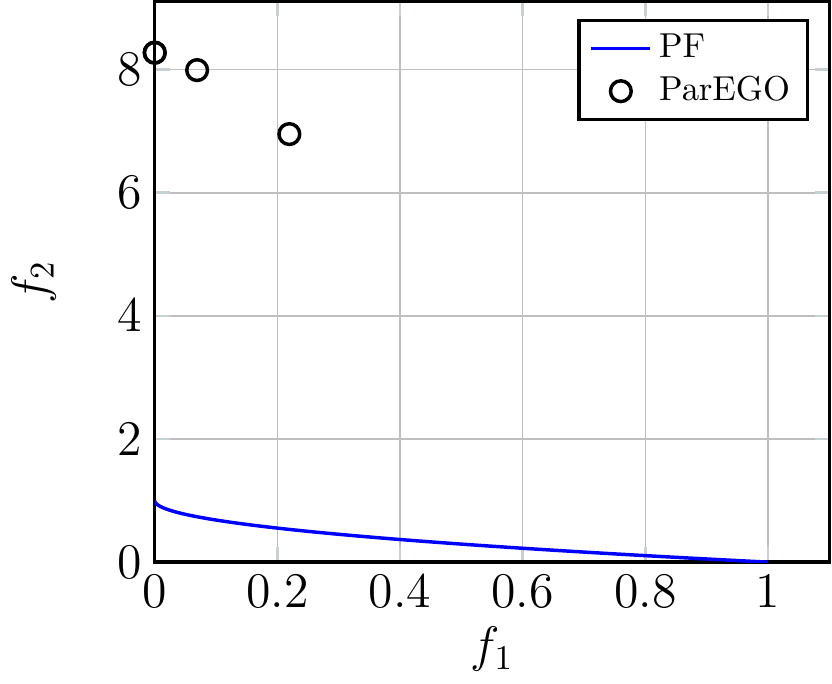}\quad
    \includegraphics[width=.22\textwidth]{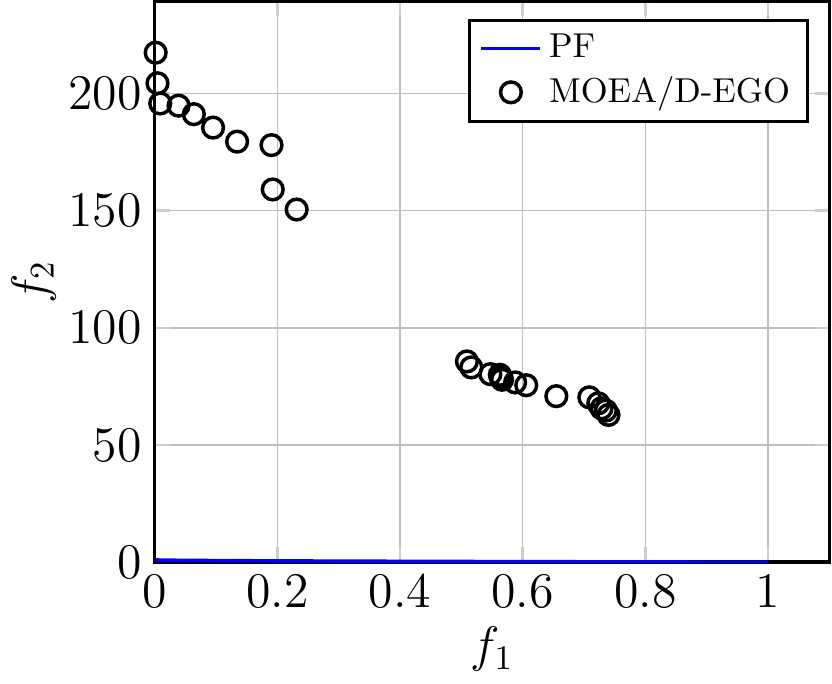}\quad
    \includegraphics[width=.22\textwidth]{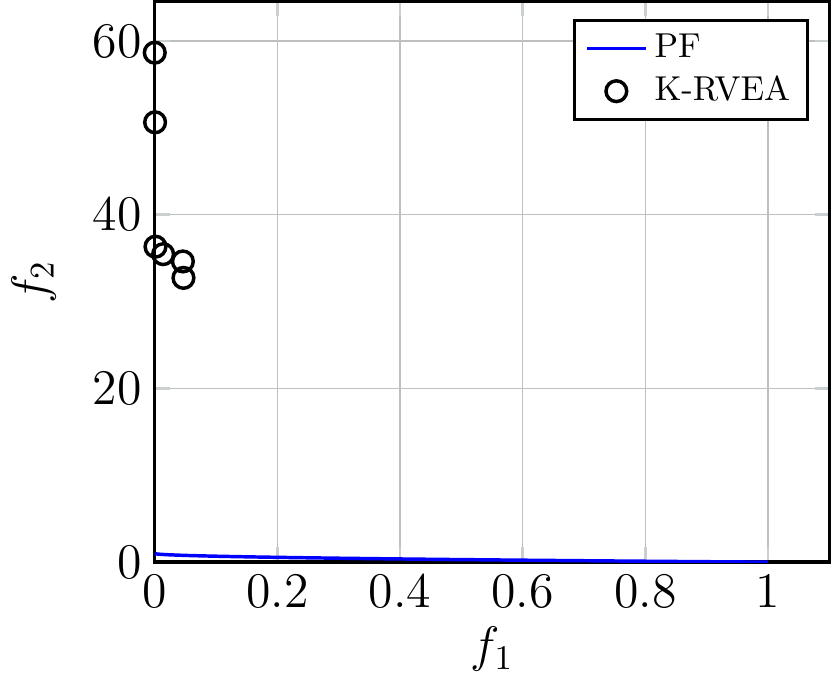}\quad
    \includegraphics[width=.22\textwidth]{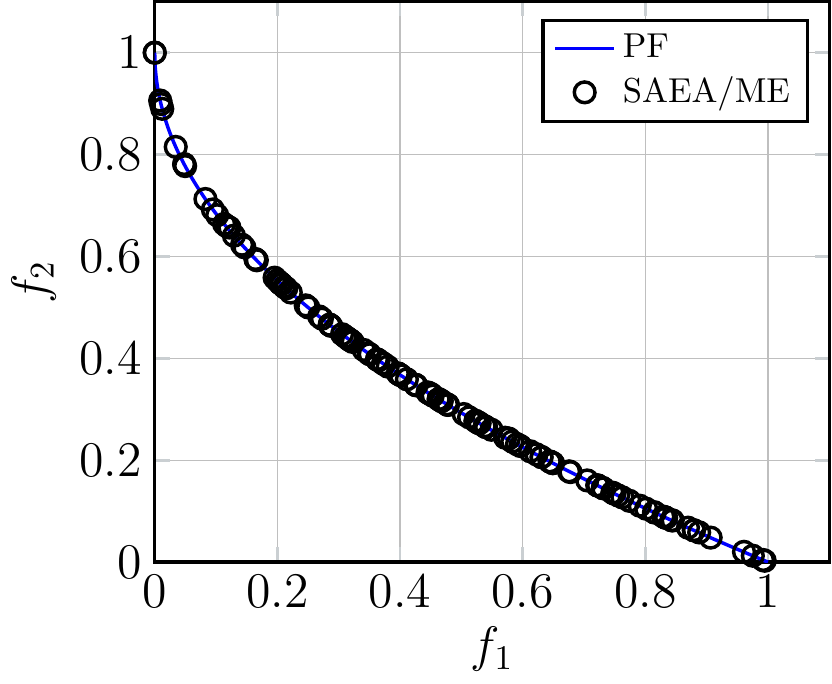}
    \caption{Non-dominated solutions obtained by four algorithms on ZDT1 ($n=50$) with the best IGD value.}
    \label{fig:ZDT1_50D}
\end{figure*}

\begin{figure*}
    \includegraphics[width=.22\textwidth]{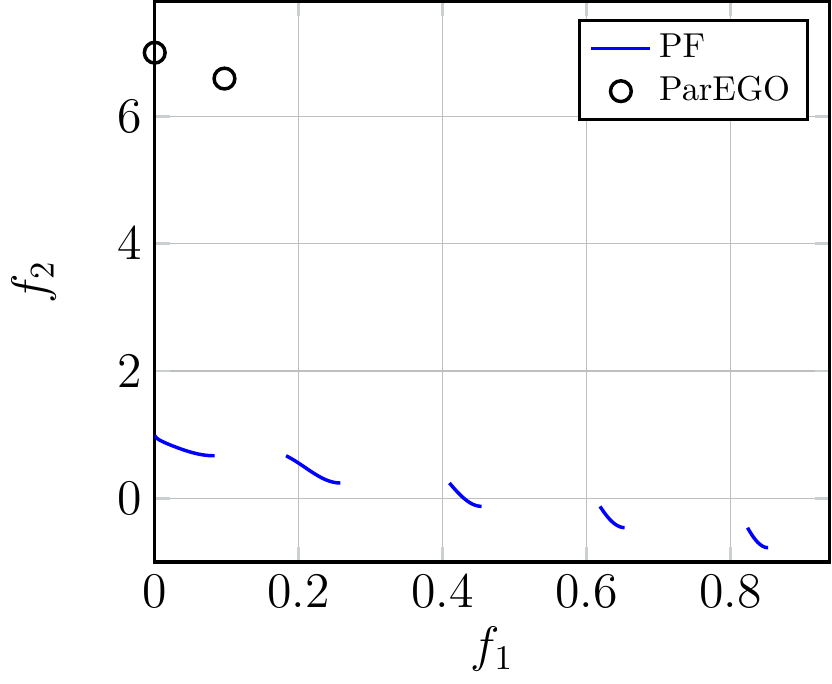}\quad
    \includegraphics[width=.22\textwidth]{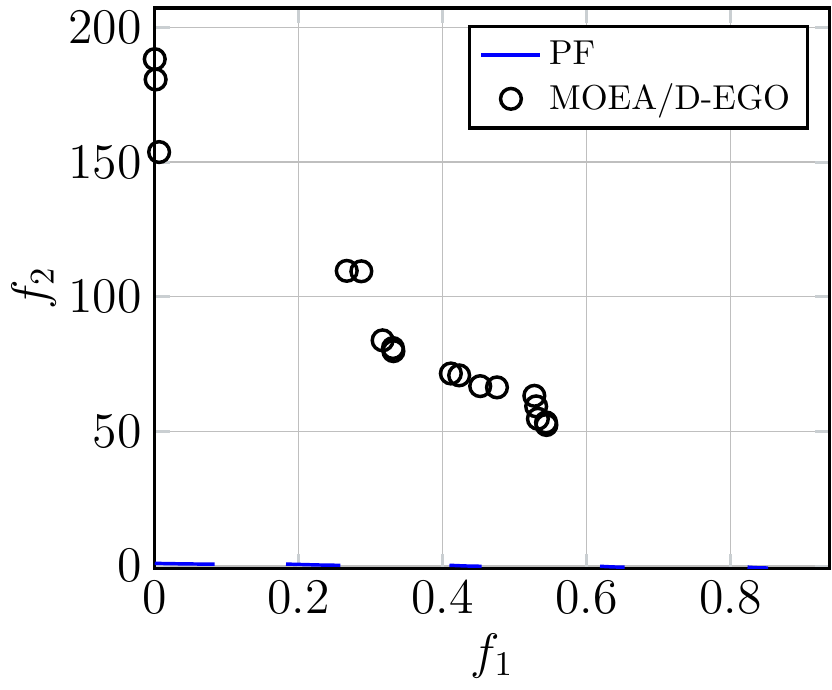}\quad
    \includegraphics[width=.22\textwidth]{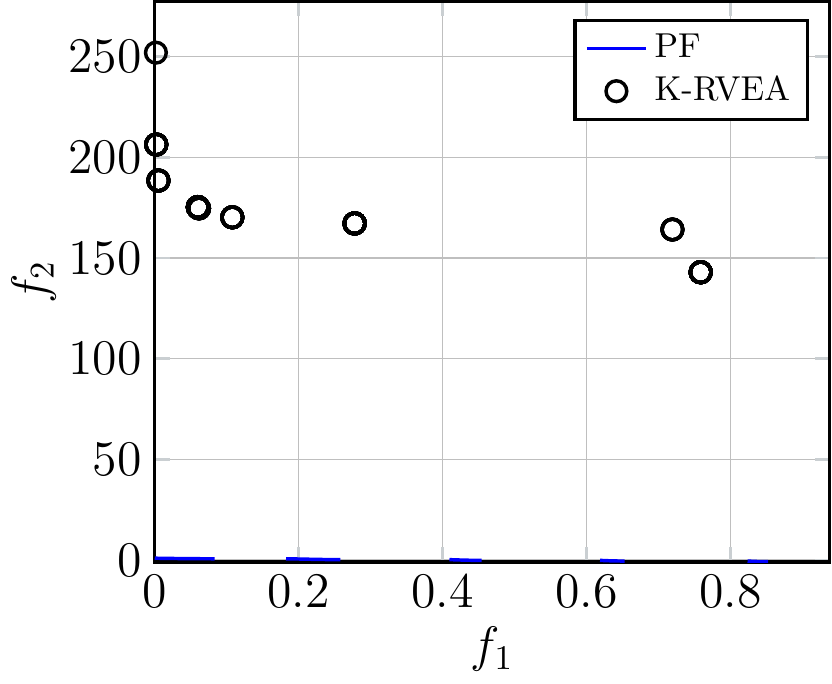}\quad
    \includegraphics[width=.22\textwidth]{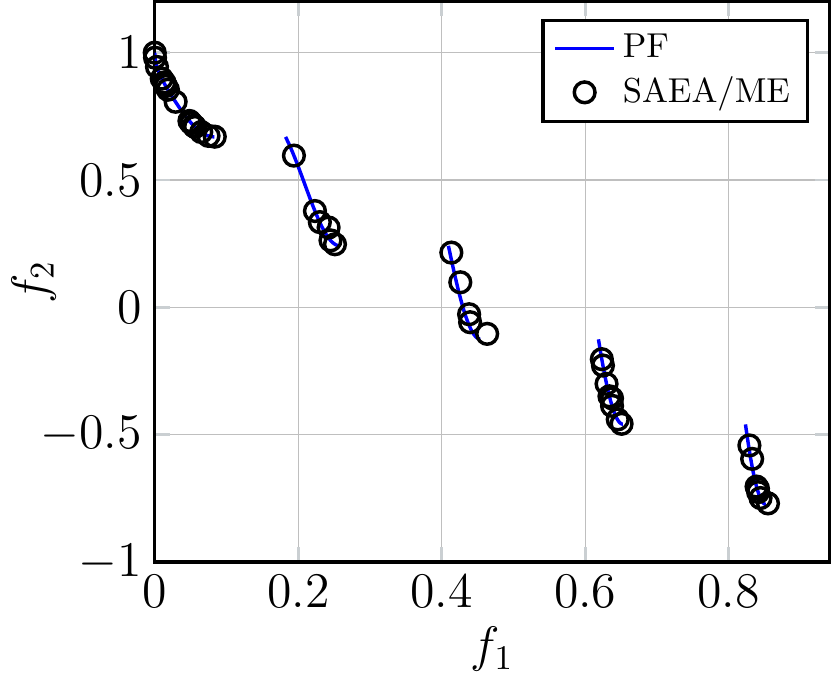}
    \caption{Non-dominated solutions obtained by four algorithms on ZDT3 ($n=50$) with the best IGD value.}
    \label{fig:ZDT3_50D}
\end{figure*}

\begin{figure*}
    \includegraphics[width=.22\textwidth]{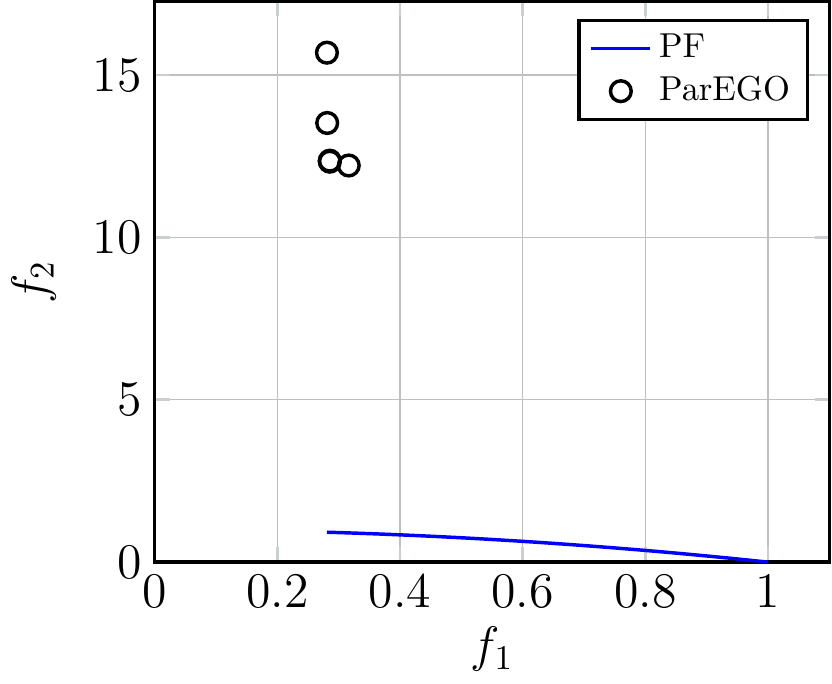}\quad
    \includegraphics[width=.22\textwidth]{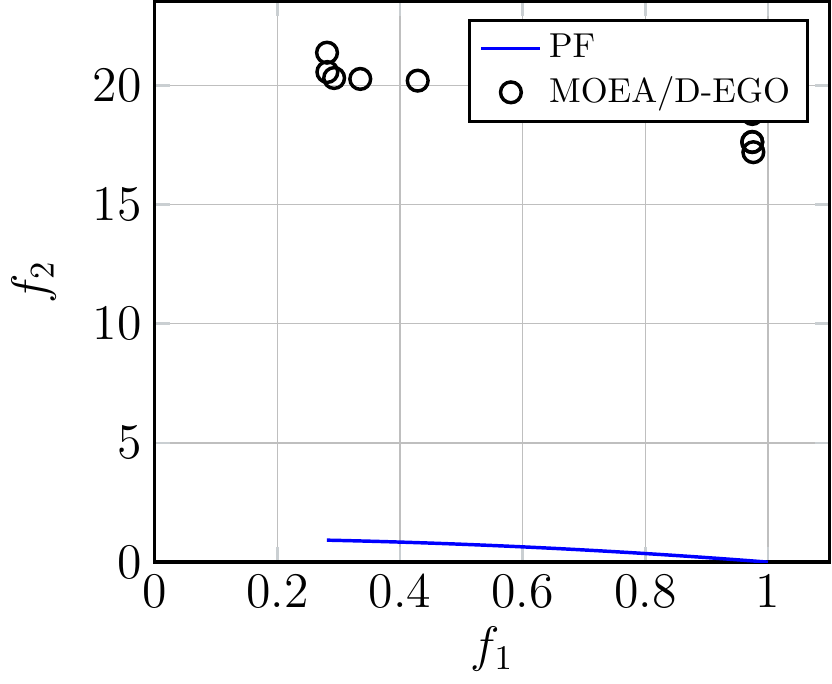}\quad
    \includegraphics[width=.22\textwidth]{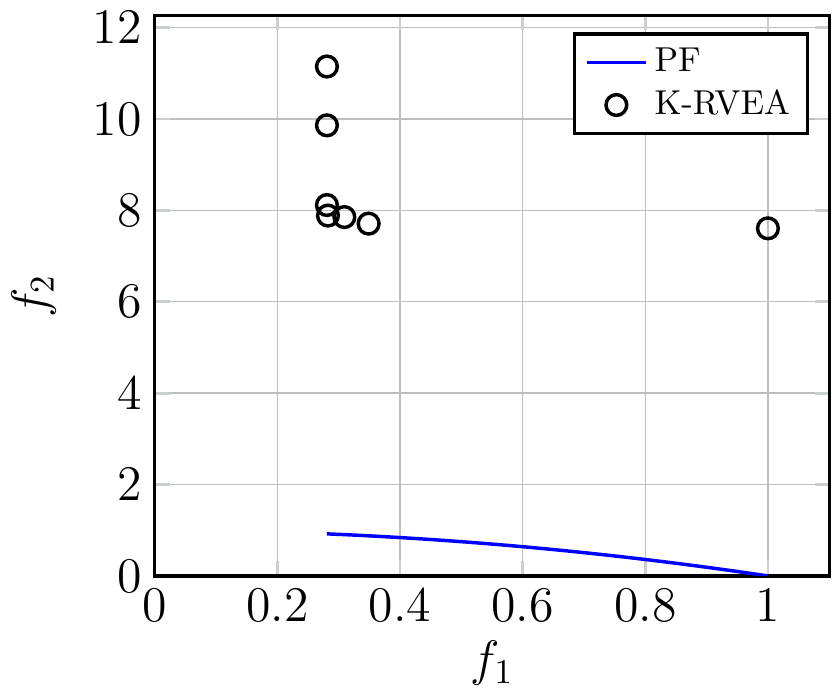}\quad
    \includegraphics[width=.22\textwidth]{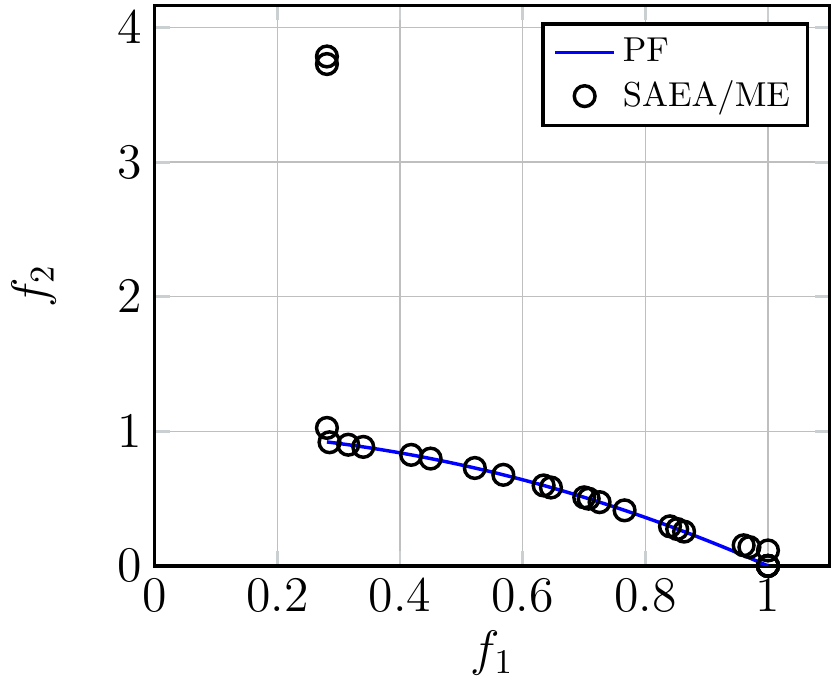}
    \caption{Non-dominated solutions obtained by four algorithms on ZDT6 ($n=50$) with the best IGD value.}
    \label{fig:ZDT6_50D}
\end{figure*}

\begin{figure*}
    \includegraphics[width=.22\textwidth]{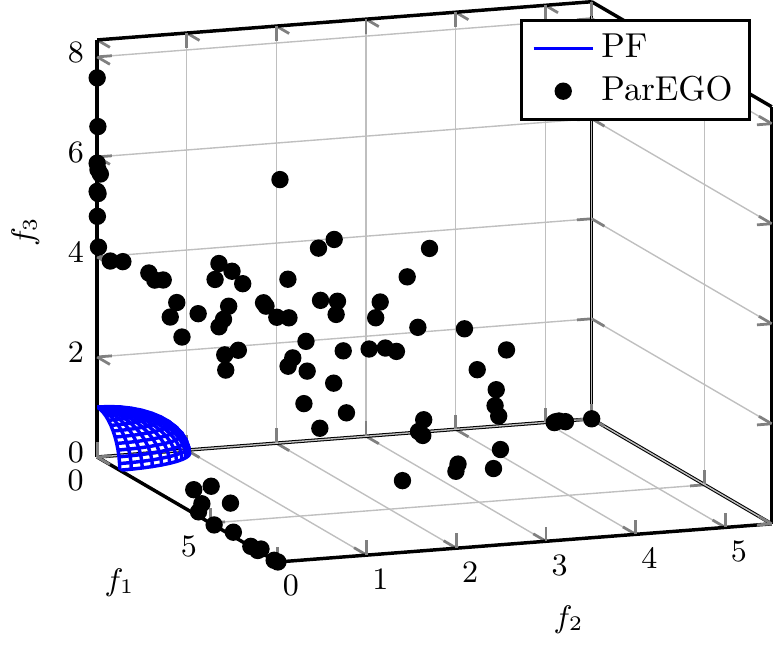}\quad
    \includegraphics[width=.22\textwidth]{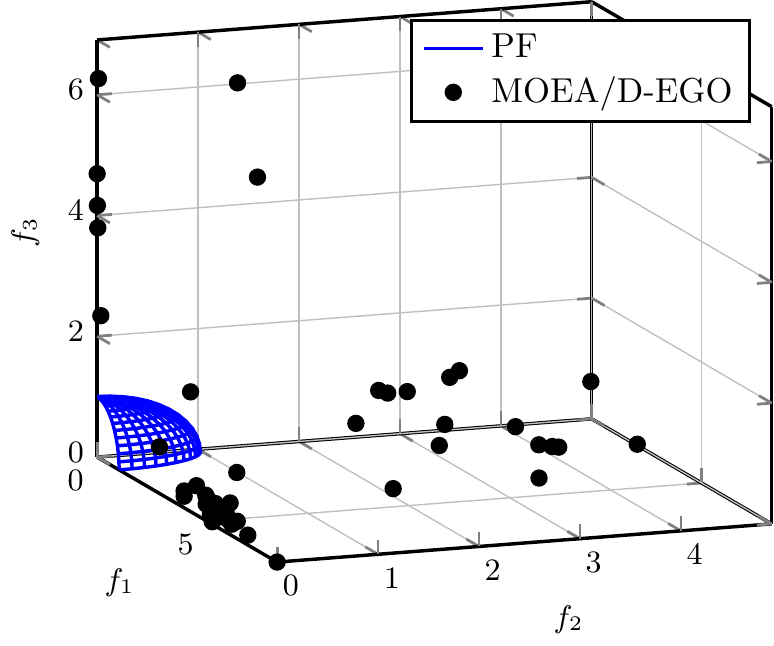}\quad
    \includegraphics[width=.22\textwidth]{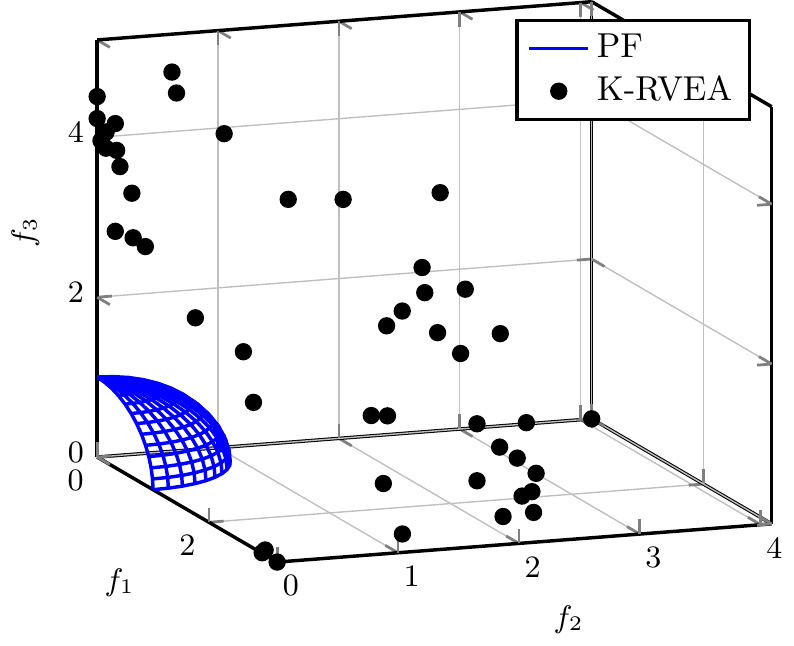}\quad
    \includegraphics[width=.22\textwidth]{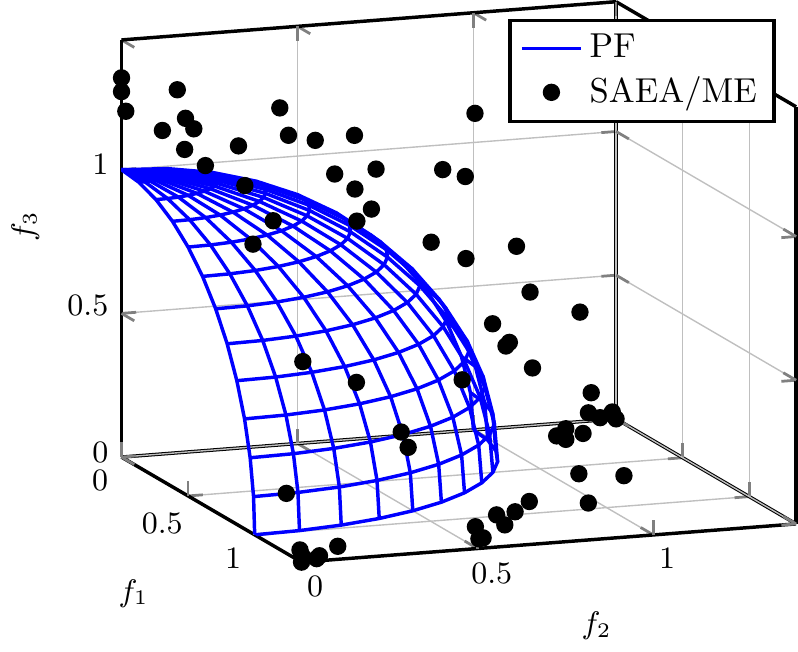}
    \caption{Non-dominated solutions obtained by four algorithms on DTLZ2 ($n=50$) with the best IGD value.}
    \label{fig:DTLZ2_50D}
\end{figure*}

\begin{figure*}
    \includegraphics[width=.22\textwidth]{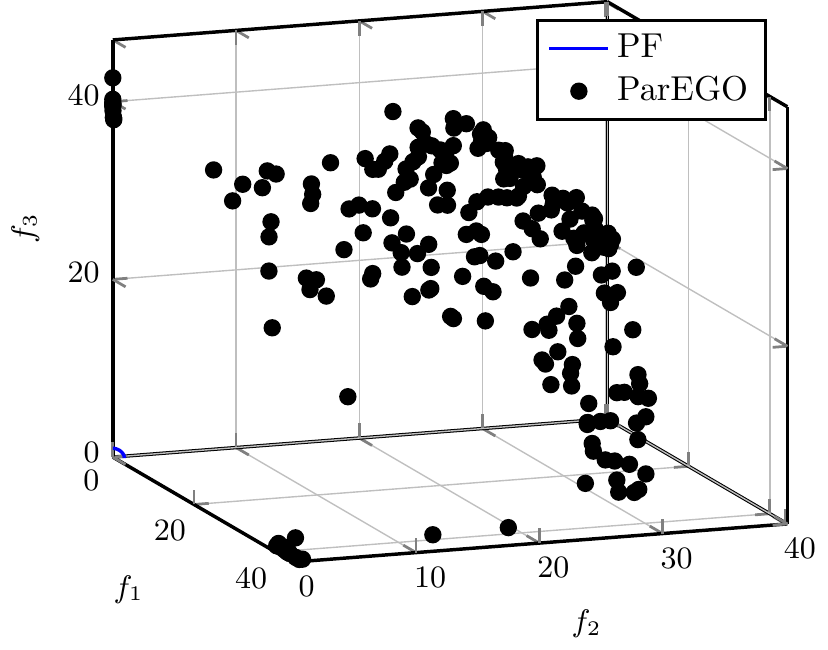}\quad
    \includegraphics[width=.22\textwidth]{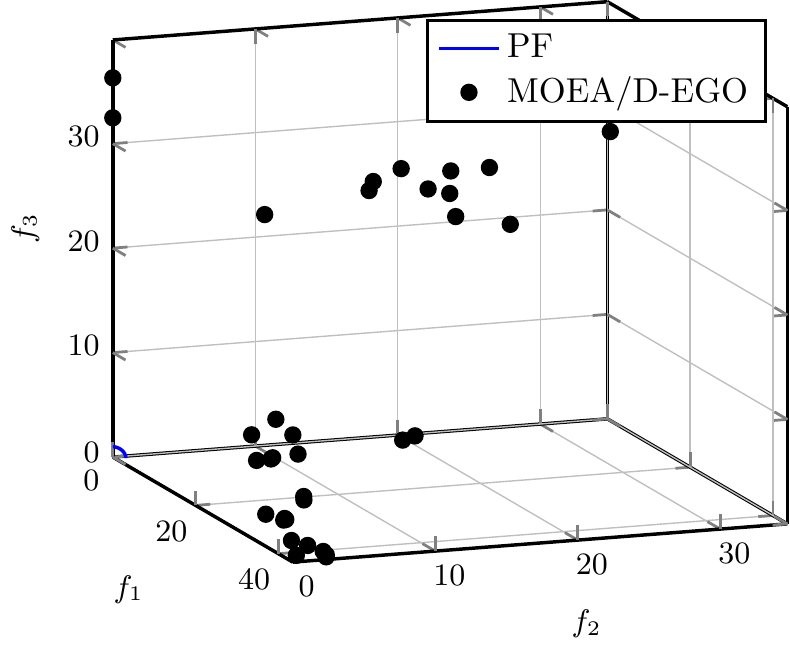}\quad
    \includegraphics[width=.22\textwidth]{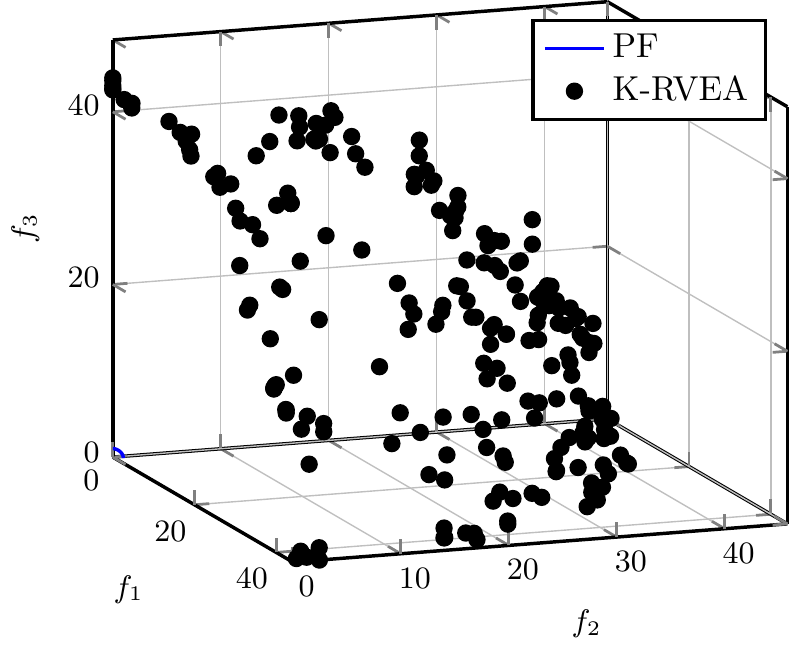}\quad
    \includegraphics[width=.22\textwidth]{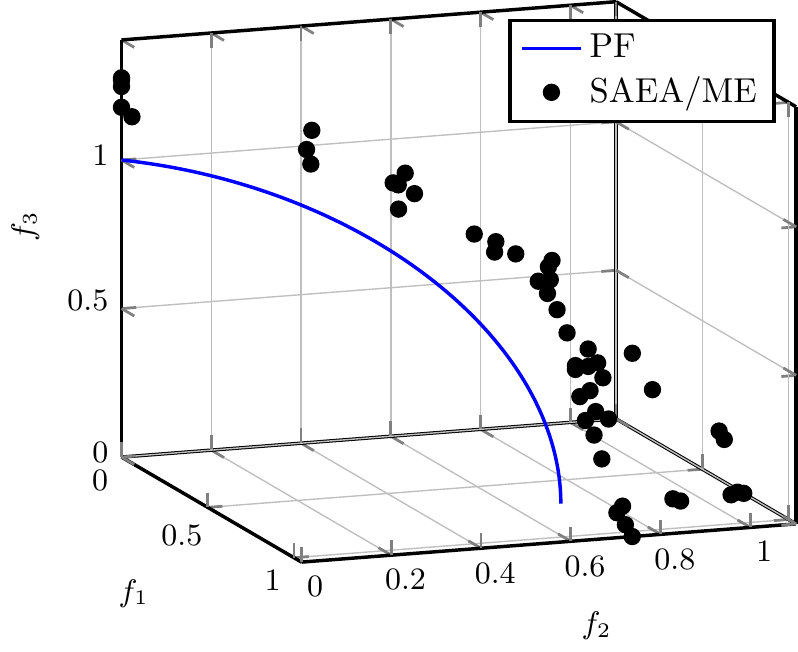}
    \caption{Non-dominated solutions obtained by four algorithms on DTLZ6 ($n=50$) with the best IGD value.}
    \label{fig:DTLZ6_50D}
\end{figure*}

\begin{figure*}
    \includegraphics[width=.22\textwidth]{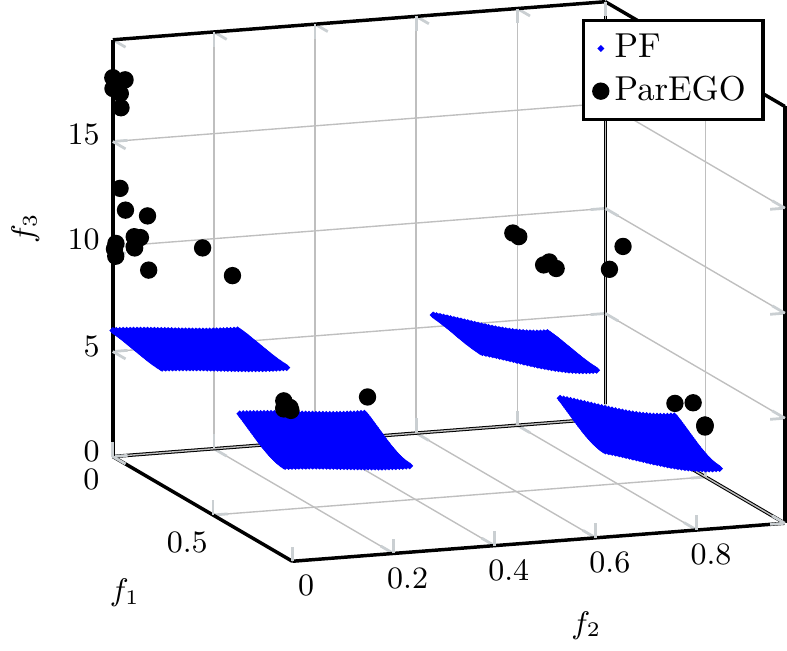}\quad
    \includegraphics[width=.22\textwidth]{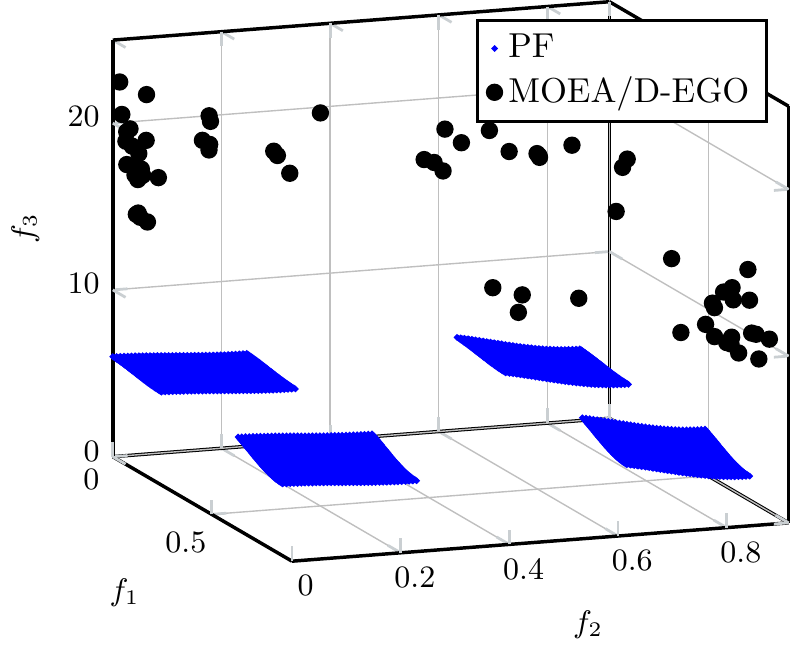}\quad
    \includegraphics[width=.22\textwidth]{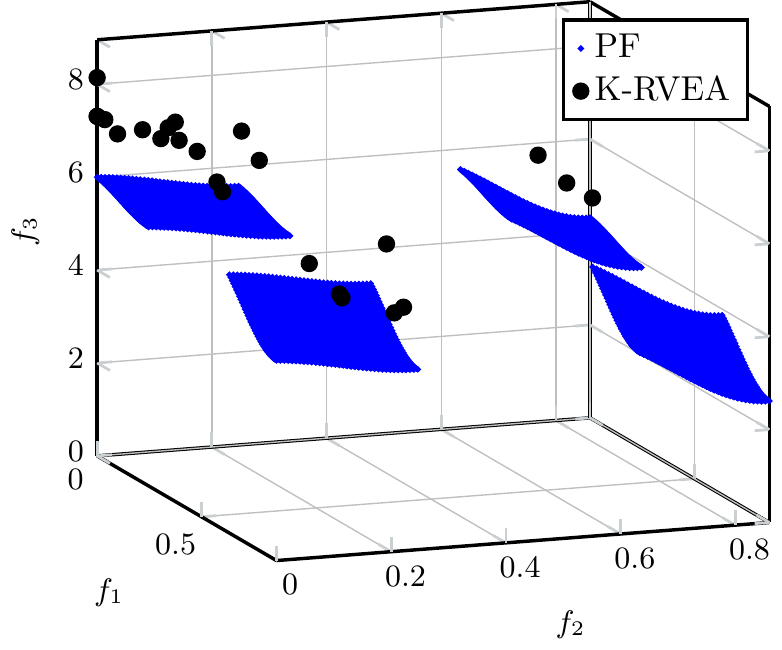}\quad
    \includegraphics[width=.22\textwidth]{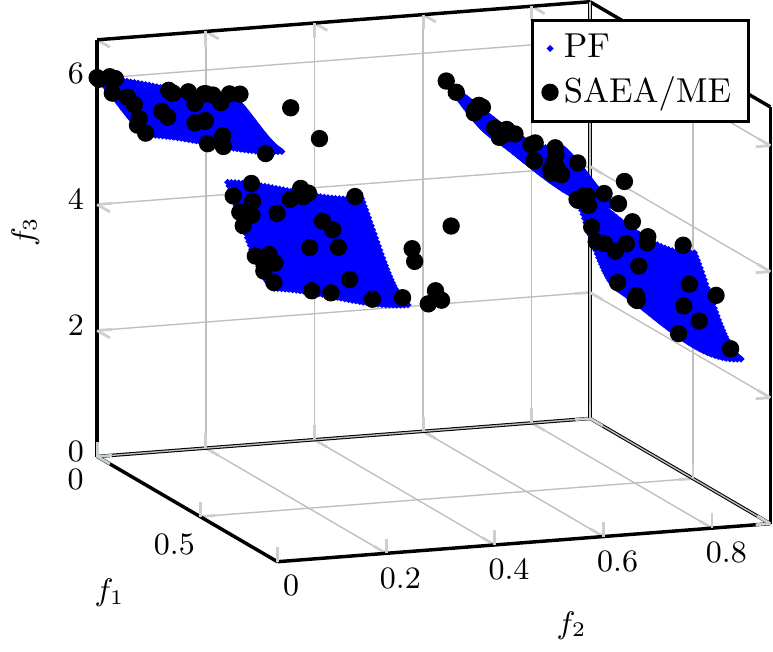}
    \caption{Non-dominated solutions obtained by four algorithms on DTLZ7 ($n=50$) with the best IGD value.}
    \label{fig:DTLZ7_50D}
\end{figure*}

\begin{figure*}
    \includegraphics[width=.22\textwidth]{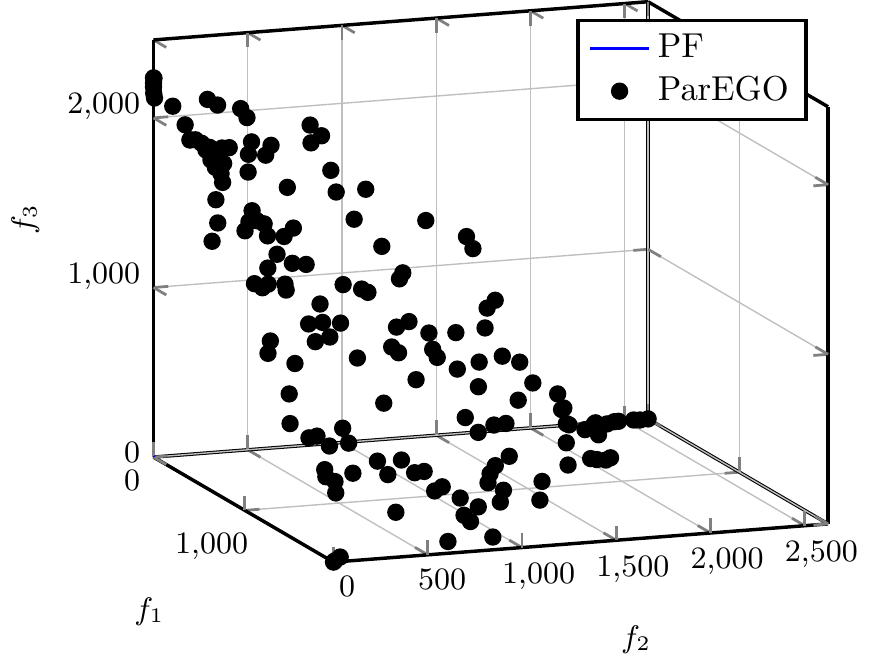}\quad
    \includegraphics[width=.22\textwidth]{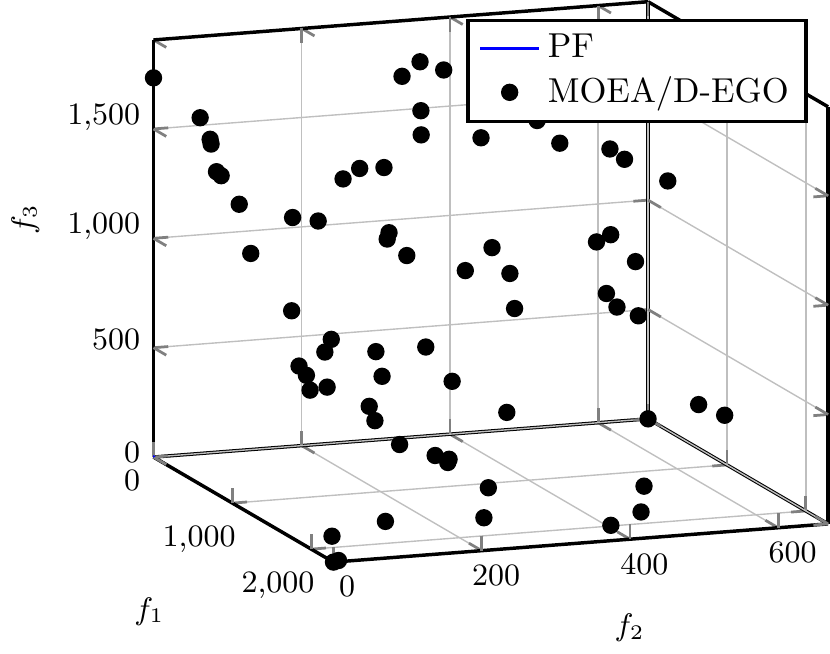}\quad
    \includegraphics[width=.22\textwidth]{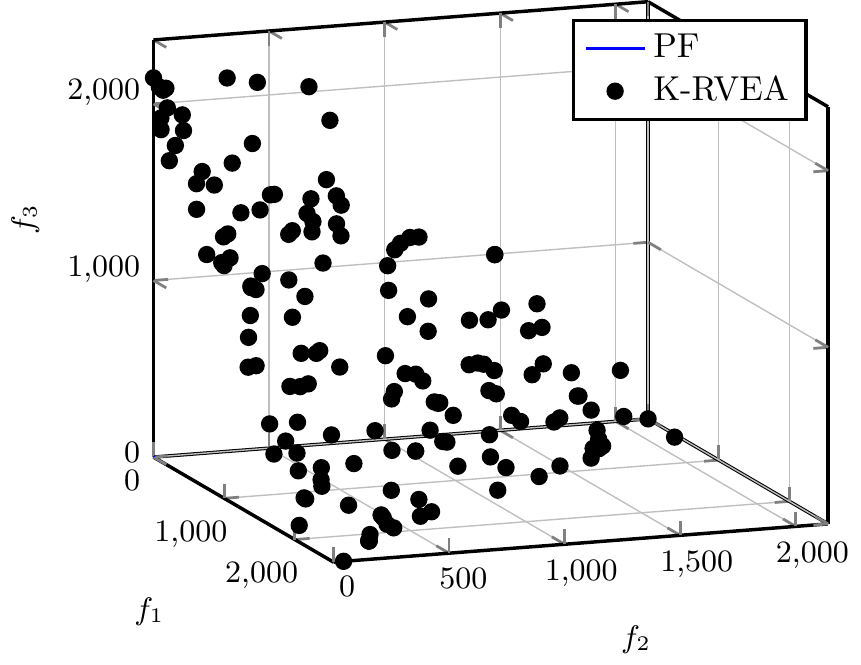}\quad
    \includegraphics[width=.22\textwidth]{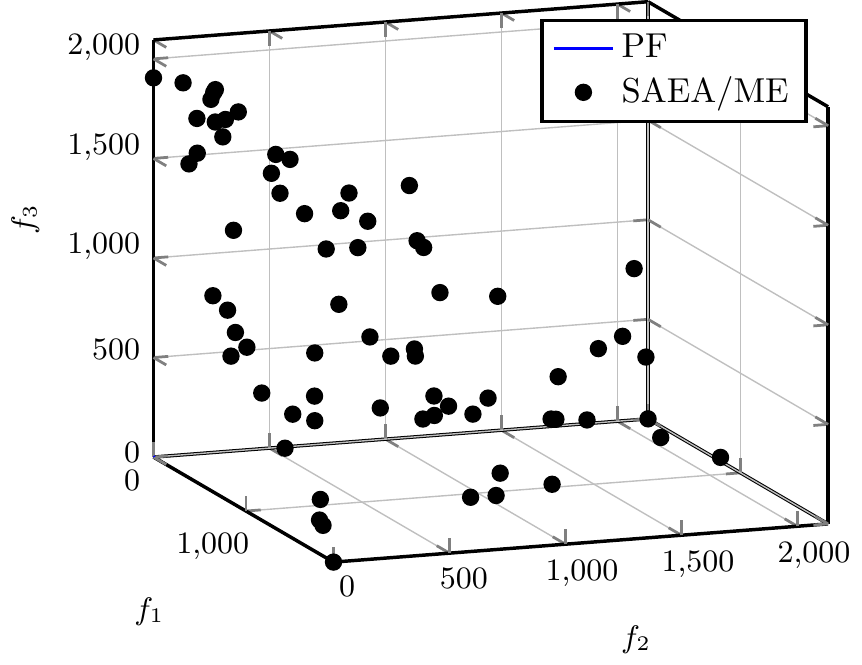}
    \caption{Non-dominated solutions obtained by four algorithms on DTLZ1 ($n=50$) with the best IGD value.}
    \label{fig:DTLZ1_50D}
\end{figure*}

In order to validate the effectiveness of SAEA/ME, this section presents the comparison results of SAEA/ME against three state-of-the-art SAEAs (i.e., ParEGO~\cite{Knowles06}, MOEA/D-EGO~\cite{ZhangLTV10} and K-RVEA~\cite{ChughJMHS18}) for expensive multi-objective optimisation. The inverted generational distance (IGD)~\cite{BosmanT03} is used as the performance metric to evaluate the performance of different algorithms. Furthermore, we choose 12 test problems from the widely used 2-objective ZDT~\cite{ZitzlerDT00} and 3-objective DTLZ~\cite{DebTLZ05} benchmark suites. Followings are some parameter settings.
\begin{itemize}
    \item The number of decision variables for each test problem considered in our experiments are 10, 20 and 50.
    \item The population size is set to 50, 100 and 300 for $n=10$, 20 and 50, respectively.
    \item The number of function evaluations is set to 300, 400 and 800 for $n=10$, 20 and 50, respectively.
	\item Each experiment is repeated 20 times and we apply the widely used Wilcoxon's rank sum test at 5\% significance level to validate the statistical significance of the results~\cite{LiZZL09,LiZLZL09,LiFK11,LiKWCR12,CaoKWL12,LiKCLZS12,LiKWTM13,LiWKC13,LiK14,CaoKWL14,CaoKWLLK15,WuKZLWL15,LiKZD15,LiKD15,LiDZ15,LiDZZ17,WuKJLZ17,WuLKZZ17,LiDY18,ChenLY18,WuLKZ20,ChenLBY18,KumarBCLB18,LiCFY19,WuLKZZ19,LiCSY19,Li19,BillingsleyLMMG19,ZouJYZZL19,LiuLC19,LiXT19,GaoNL19}.
	\item In the subset selection for model management, we set $k=10$ in our experiments.
\end{itemize}

\subsection{Experimental Results}
\label{sec:experiments}

The comparison results of IGD values obtained by different algorithms are given in~\pref{tab:results}. In particular, the best result for each test problem instance is highlighted in bold face with a gray background. From these comparison results, we can clearly see the overwhelmingly better performance achieved by SAEA/ME in 95 out of 108 comparisons. In the following paragraphs, we will give a gentle discussion over these results.

ZDT1 and ZDT2 are relatively simple test problems, on which all four algorithms do not have too much difficulty to converge to the global PFs when the number of variables is small (i.e., $n=10$), as shown in Figure 1 and Figure 4 in the supplementary document\footnote{Due to the space limit, the comprehensive results of population plots can be found from \url{https://tinyurl.com/yx4pmwwq}}. However, the performance of MOEA/D-EGO and K-RVEA degenerate dramatically when having more variables. ParEGO performs slightly better than MOEA/D-EGO and K-RVEA when $n=20$, but its IGD becomes over 10 which indicates its poor convergence. As shown in~\pref{fig:ZDT1_50D}, none of ParEGO, MOEA/D-EGO and K-RVEA can find any solution on the PF. In contrast, the performance of SAEA/ME is relatively consistent across $n=10$ to $n=50$. 

ZDT3 is a discontinuous problem whose PF is five disconnected segments. Similar to the observations on ZDT1 and ZDT2, all algorithms do not have too much trouble to find solutions on the PF when $n=10$, but their performance degenerate with the increase of the number of variables except SAEA/ME. In particular, it is interesting to see that ParEGO and K-RVEA can still find meaningful solutions when $n=20$ whereas MOEA/D-EGO can hardly converge even in this case.

The PF of ZDT6 has a biased distribution. All four algorithms cannot find fully converged solutions on this problem even when $n=10$. However, most solutions found by SAEA/ME are along the PF whereas those found by the other three algorithms are way beyond the PF. It is also interesting to see that the performance of SAEA/ME on problems with 50 variables is even better than those with less variables according to~\pref{tab:results}.

As for the 3-objective DTLZ test problems, the performance of SAEA/ME also degenerate with the increase of the number of variables. Nevertheless, such degeneration is not as significant as the other three peer algorithms. More specifically, DTLZ2 is a relatively simple test problem. But due to the increase of number of objectives, the solutions found by SAEA/ME are not well converged. In contrast, ParEGO and MOEA/D-EGO can hardly find a converged solution even on DTLZ2. DTLZ4 has the same same PF shape as DTLZ2, but it has a strong bias which makes algorithms difficult to find a set of well distributed solutions. As shown in~\pref{tab:results}, K-RVEA shows better performance than SAEA/ME when $n=10$ and 20. DTLZ5 and DTLZ6 are MOPs with a degenerated PF. From the results shown in~\pref{fig:DTLZ6_50D}, we can see that only SAEA/ME can find solutions close to the PF whist the solutions found by the other three algorithms are way beyond the PF. DTLZ7 is a test problem with disconnected PF segments. K-RVEA is the best algorithm when $n=20$. However, it is interesting to see that the performance of SAEA/ME become better when having a larger number of variables. As shown in~\pref{fig:DTLZ7_50D}, only SAEA/ME can find solutions on the PF whereas the solutions found by the other three algorithms are away from the PF.

Although the performance of SAEA/ME is satisfactory in most cases, it failed to find any meaningful solution for ZDT4, DTLZ1 and DTLZ3. In particular, these three test problems are with many local optima. As the results shown in~\pref{tab:results}, the IGD values obtained by all four algorithms are over 10 across all test instances. As the population plots shown in~\pref{fig:DTLZ1_50D}, we can clearly see that solutions obtained by all algorithms are way beyond the PF. The poor performance of all these algorithms on ZDT4, DTLZ1 and DTLZ3 can be attributed to their multi-modality which makes the surrogate modelling become even more difficult.

%% file: conclusion.tex

\section{Conclusions}
\label{sec:conclusion}

Building a surrogate model of the originally computationally expensive objective function has been recognised as the stepping stone of EA towards a wider range of application in the real world. However, due to the curse-of-dimensionality, most existing research on SAEAs have been wandered in problems with a relatively small number of variables. To address the scalability issue, this paper proposed SAEA/ME for solving medium-scale MOPs with less than 50 variables. According to the experimental results, we have witnessed a clear superiorly of our proposed SAEA/ME against three state-of-the-art SAEAs in over 85\% comparisons. The success of SAEA/ME can be attributed to three distinctive features.
\begin{itemize}
\item To combat the curse-of-dimensionality in surrogate model building, the surrogate models in SAEA/ME are built upon a reduced feature space by analysing the correlation relationship between decision variables and objective functions.
\item To strike a better balance between exploration and exploitation, the underlying MOP is transformed to a many-objective optimisation problem formulation based on the surrogate objective functions and their associated estimations of uncertainty.
\item To implement a model management in a batch manner, a subset selection method is proposed to select a couple of promising solutions for actual function evaluations.
\end{itemize}

Although SAEA/ME is only tested on medium-scale MOPs, it does not mean that SAEA/ME is not scalable any further. In future, it is interesting to investigate other dimensionality reduction techniques to mitigate the curse-of-dimensionality in surrogate model building. It is arguable to use NSGA-II to optimise a many-objective optimisation problem in the model-based search process, given its reported drawbacks for many-objective optimisation~\cite{IshibuchiISN16}. We will investigate other many-objective optimiser in this optimisation.